\documentclass[fleqn,10pt]{wlscirep}
\graphicspath{{figures/}}
\usepackage[utf8]{inputenc}
\usepackage[T1]{fontenc}
\usepackage{rotating}
\usepackage{multirow}
\usepackage{threeparttable}
\usepackage{hyperref}
\usepackage{appendix}
\usepackage{float}
\usepackage{xcolor}
\usepackage{booktabs}
\usepackage{tabularx}
\usepackage{colortbl}
\usepackage{listings}
\usepackage{xurl} 

\newcommand{\filepath}[1]{\nolinkurl{#1}}
\newcommand{\col}[1]{\nolinkurl{#1}}
\lstdefinestyle{walkthrough}{
  basicstyle=\ttfamily\scriptsize,
  breaklines=true,
  breakatwhitespace=true,
  columns=fullflexible,
  keepspaces=true,
  showstringspaces=false,
  frame=single,
  framesep=2pt,
  xleftmargin=4pt,
  xrightmargin=4pt,
  upquote=true,
  literate={’}{{'}}1 {‘}{{`}}1 {“}{{"}}1 {”}{{"}}1 {—}{{---}}1 {–}{{--}}1
}
\definecolor{krheader}{RGB}{230,236,245}
\definecolor{krours}{RGB}{236,243,253}
\newcommand{\figref}[1]{Figure~\ref{#1}}
\newcommand{\tabref}[1]{Table~\ref{#1}}

\newcommand{\appref}[1]{Appendix~\ref{#1}}

\title{A knowledge-augmented dataset of high-risk driving scenarios with LLM annotations for autonomous driving}

\author[1]{Heye Huang}
\author[1]{Jingguang Li}
\author[2,*]{Zhiyuan Zhou}
\author[3]{Paul Liang}
\author[4]{Mingyu Wu}
\author[1]{Kitae Jang}
\author[5,*]{Jianqiang Wang}

\affil[1]{\small Cho Chun Shik Graduate School of Mobility, Korea Advanced Institute of Science and Technology, Daejeon 34051, South Korea}
\affil[2]{\small College of Computer Science and Artificial Intelligence, Fudan University, Shanghai 200433, China}
\affil[3]{\small Media Lab, Massachusetts Institute of Technology, Cambridge, MA 02139, USA}
\affil[4]{\small School of Mechanical Engineering, Shanghai Jiao Tong University, Shanghai 200240, China}
\affil[5]{\small School of Vehicle and Mobility, Tsinghua University, Beijing 100084, China}

\affil[*]{Correspondence (zhouzhiyuan@pjlab.org.cn, wjqlws@tsinghua.edu.cn).}

\begin{abstract}
Safe autonomous driving requires both a rapid response to common high-risk events and deeper reasoning over the rare, extreme long-tail scenarios of traffic safety. These scenarios are severely under-represented in naturalistic driving data, and existing trajectory and language-augmented datasets seldom provide high-risk event labels, semantic annotations and verifiable safety signals. Here we present K-Risk, a knowledge-augmented dataset that combines structured driving trajectories with large language model (LLM)-generated semantic annotations for safety-critical driving scenarios. K-Risk integrates 20 human-driven and autonomous-vehicle trajectory datasets from Europe, China, and the United States, covering highways, urban freeways, intersections and roundabouts. Using a unified risk-centric extraction pipeline, K-Risk curates 31,398 high-risk events, together with a 1,036-event extreme subset of near-collision cases. Each event is released as a synchronized trajectory–metadata–language triplet containing structured scenario descriptions, abnormal-behavior notifications, and, for a representative subset, LLM-generated causal risk analyses and action recommendations validated through a closed-loop simulator with iterative reflection. By combining multi-dimensional risk annotations, interpretable language supervision, and verifiable decisions, K-Risk bridges structured traffic trajectories, semantic reasoning and verifiable decision supervision, providing a standardized foundation for developing and evaluating next-generation risk-aware autonomous driving agents.
\end{abstract}

\begin{document}

\flushbottom
\maketitle

\thispagestyle{empty}
\section*{Background \& Summary}

Autonomous driving has progressed rapidly from partial to conditional automation, yet safety in rare and complex long-tail scenarios remains the central barrier to large-scale deployment and on-road testing~\cite{sun2020scalability}. A safe driving agent must combine two capabilities that draw on different kinds of experience: a fast and reliable response to the common high-risk events encountered every day, and a slower, deeper form of reasoning over the rare and extreme situations that sit in the long tail of traffic safety. The second capability is harder to acquire, because the corresponding scenarios occur infrequently and are therefore severely under-represented in the naturalistic driving data used to train and evaluate driving models. Reliable behavior in these high-risk edge cases is decisive not only for technical performance but also for public trust and regulatory approval~\cite{wilson2023argoverse}. Agents based on large language models (LLMs) have recently emerged as a promising route to this second capability. By combining structured driving knowledge, learned experience and semantic understanding, they offer a interpretable reasoning process that interprets a scenario, reflects on prior cases and generalizes across situations. They can also be coupled with physics-based models to yield systems that are at once safer and more interpretable~\cite{han2025dme, zhou2024safedrive}. Realizing this potential, however, depends on training and evaluation data that pair real high-risk trajectories with the semantic, interpretable and verifiable signals these agents consume, and such data remain scarce.

Naturalistic trajectory and perception datasets such as highD~\cite{krajewski2018highd}, inD~\cite{bock2020ind}, Waymo~\cite{sun2020waymo}, nuScenes~\cite{caesar2020nuscenes} and Argoverse~\cite{chang2019argoverse} have substantially advanced perception, tracking and trajectory forecasting by providing precise trajectories, multimodal sensor streams and standardized benchmarks. Their records nonetheless capture mostly routine driving in structured environments, so that genuine edge cases and high-risk interactions make up only a small fraction of the data. This composition limits their value for training and evaluating autonomous driving systems under safety-critical conditions.
Datasets that focus on interaction narrow this gap only partially. OnSiteVRU~\cite{yan2025onsitevru} emphasizes dense urban scenarios with vulnerable road users, yet the most unpredictable behaviors, such as sudden pedestrian dart-outs, remain underrepresented; INTERACTION~\cite{zhan2019interaction} covers roundabouts, merges and near-collision events with semantic maps; and InterHub~\cite{jiang2024interhub} formalizes dense multi-agent interaction through the post-encroachment time and the minimum sum of absolute acceleration. Across these datasets, risk is still expressed mainly through one-dimensional proximity metrics such as time-to-collision or time headway, which oversimplify danger and fail to reflect how human drivers read a situation through multi-agent dynamics, intent and context~\cite{mokhtarian2022time}.

A second line of datasets adds semantic, natural language annotation to support the higher-level reasoning that LLM-based agents require. CoVLA~\cite{arai2024covla} pairs real-world driving video with detailed descriptions of environments and maneuvers; OmniDrive~\cite{wang2025omnidrive} provides holistic vision-language annotations and counterfactual question answering; DriveLM~\cite{sima2024drivelm} structures perception, prediction and planning as graph-based question answering; and CODA-LM~\cite{li2024codalm} targets corner-case understanding through a categorical taxonomy. More recent efforts move toward risk and closed-loop evaluation: NuRisk~\cite{gao2025nurisk} builds an agent-level risk visual-question-answering benchmark with quantitative risk scores over bird's-eye-view sequences, and Impromptu VLA~\cite{chi2025impromptuvla} distills unstructured corner-case clips for vision-language-action fine-tuning. These datasets are rich in language, yet they center on routine driving, annotate risk coarsely, and rarely expose the causal risk analyses or verifiable safety signals needed to train and evaluate risk-aware agents~\cite{arai2024covla, wang2025omnidrive}.

\begin{figure}[htbp]
\centering
\includegraphics[width=\linewidth]{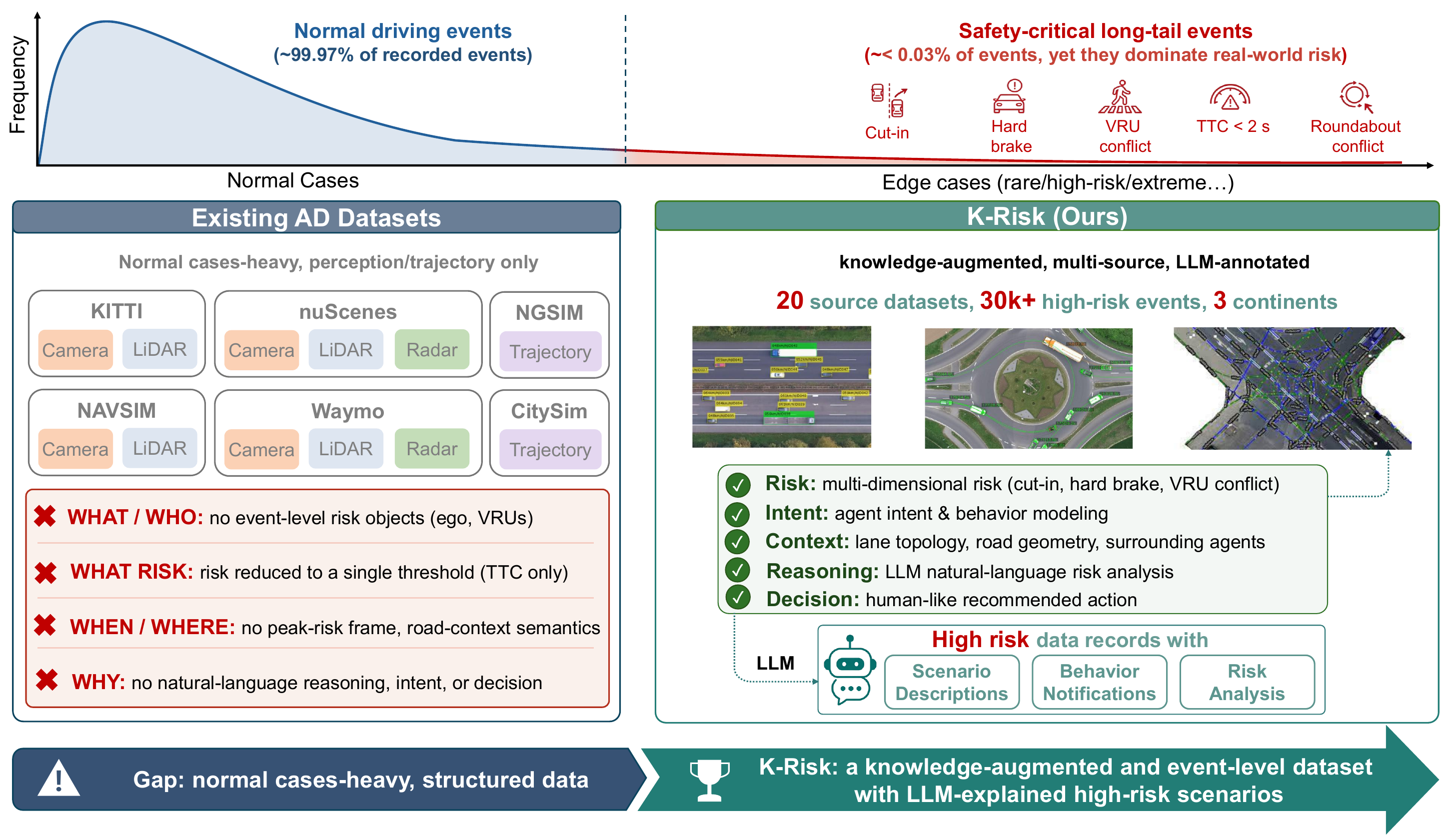}
\caption{Motivation for K-Risk. Naturalistic driving data are dominated by routine cases, whereas safety-critical long-tail events such as cut-ins, hard braking, VRU conflicts, low-TTC situations and roundabout conflicts are rare but important for autonomous driving safety. Existing datasets mainly provide perception or trajectory records for normal driving and often lack event-level risk objects, multi-dimensional risk labels, peak-risk context and natural language reasoning. K-Risk addresses these gaps by integrating 20 trajectory sources across three continents into more than 30{,}000 high-risk events, each paired with structured scenario descriptions, behavior notifications and, for a representative subset, LLM-generated risk analyses.}
\label{fig:motivation}
\end{figure}

Taken together, these datasets leave four gaps that limit the development of risk-aware driving agents, as summarized in \figref{fig:motivation}. First, safety-critical events are inherently rare and are seldom annotated with a consistent, multi-dimensional definition of risk. Second, existing datasets generally lack event-level semantic annotations that explain what occurred, why the situation became hazardous, and how the risk evolved over time. Third, although large language models have shown remarkable capability in reasoning and decision making, few driving datasets provide structured, machine-readable natural language descriptions that can be directly used for language-based learning or evaluation. Finally, existing datasets rarely provide verifiable decision signals that enable recommended actions to be validated in a closed-loop manner.

\begin{figure}[!t]
\centering
\includegraphics[width=\linewidth]{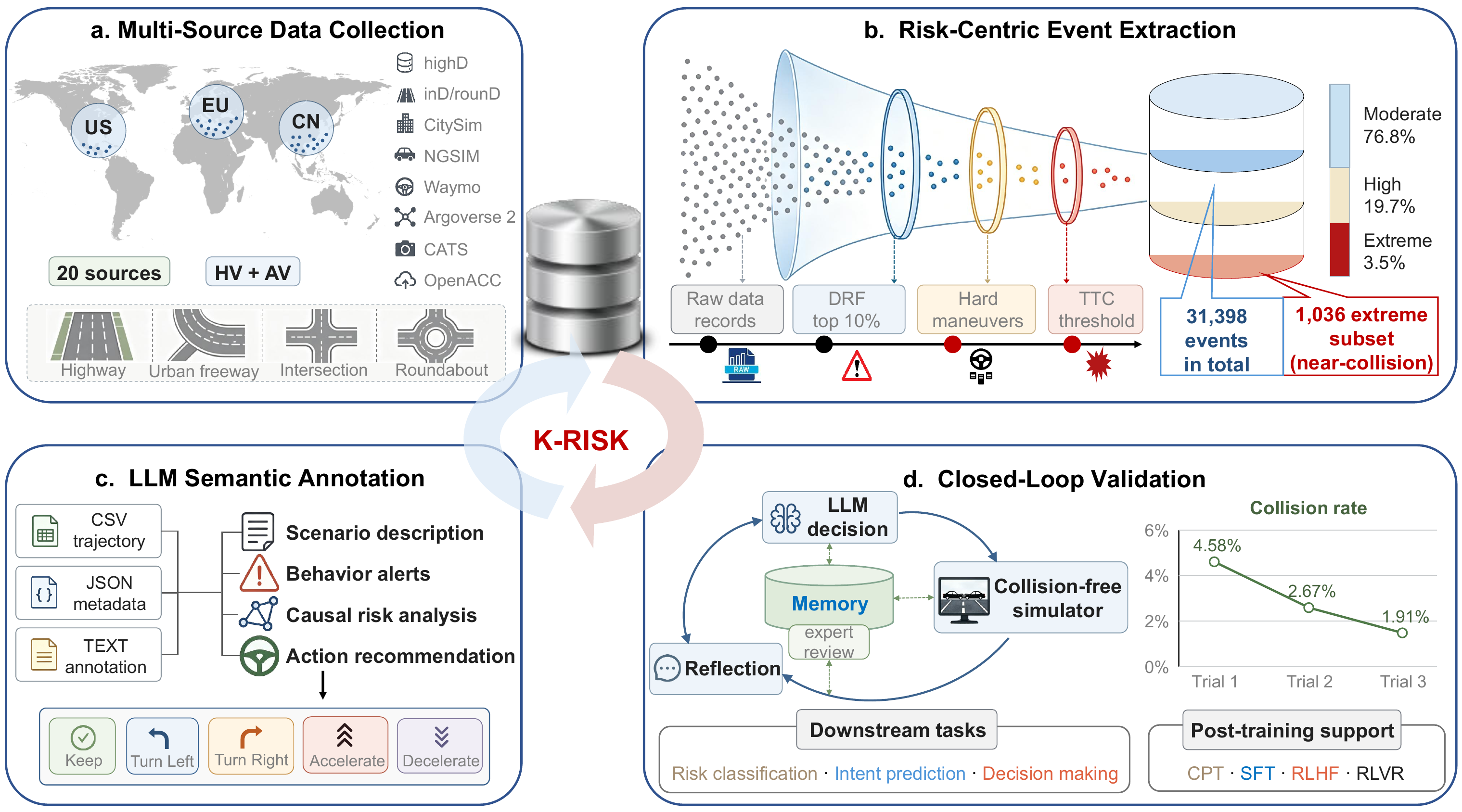}
\caption{Overview of the K-Risk dataset and the LLM annotation framework. \textbf{(a)} Multi-source curation aggregates 20 human-driven (HV) and autonomous-vehicle (AV) trajectory sources across the United States, Europe and China, spanning highways, urban freeways, intersections and roundabouts. \textbf{(b)} Risk-centric event extraction screens the raw records through a driver risk field filter, calibrated hard-maneuver thresholds and a TTC and trajectory-conflict test, yielding 31{,}398 events graded as moderate, high or extreme, with a 1{,}036-event extreme near-collision subset. \textbf{(c)} Semantic annotation converts each event into synchronized CSV, JSON and text records carrying a scenario description and behavior notifications, with an LLM causal risk analysis and an action recommendation drawn from a five-action schema added for a representative subset. \textbf{(d)} Closed-loop validation checks each recommended action against a collision-free simulator with reflection, supporting downstream tasks and the post-training stages CPT, SFT, RLHF and RLVR.}
\label{fig:framework}
\end{figure}

To address these limitations, we introduce K-Risk, a knowledge-augmented dataset of high-risk driving scenarios that links structured vehicle trajectories with event-level semantic annotations, as illustrated in \figref{fig:framework}. K-Risk is constructed from 20 trajectory sources that combine human-driven-vehicle (HV) and autonomous-vehicle (AV) data across Europe, China, and the United States. These sources cover highways, urban freeways, intersections, and roundabouts, allowing the dataset to reflect diverse traffic rules, road geometries, and driving conventions. The six HV sources are highD, inD, and rounD from the LevelXData collection, ExpresswayA and FreewayB from CitySim, and the I-80 dataset from NGSIM. The fourteen AV sources are the Argoverse~2 Motion Forecasting, Waymo Open Motion and Waymo Open Perception datasets, the MicroSimACC dataset, the CATS ACC, CATS Platoon and CATS UWM datasets, the Central Ohio single-vehicle and two-vehicle datasets, the OpenACC Casale, Vicolungo, AstaZero and ZalaZONE datasets, and the Vanderbilt ACC experiments.

From these sources, K-Risk curates 31,398 high-risk events, including 1,036 extreme near-collision cases. Events are identified using a unified protocol that combines a driver risk field filter, calibrated behavioral thresholds, and a two-second trajectory-conflict predictor. Each event is enriched with structured scenario descriptions and abnormal-behavior notifications, while a representative subset further includes LLM-generated causal risk analyses and discrete-action recommendations validated in collision-free simulation. Stored as synchronized trajectory data, metadata, and natural language annotations, K-Risk provides risk labels, semantic context, and verifiable safety signals. Key contributions are as follows:

\begin{itemize}

\item
We introduce K-Risk, a large-scale knowledge-augmented dataset of high-risk driving scenarios built from 20 public trajectory datasets across three continents. K-Risk contains 31,398 curated high-risk events, including a 1,036-event extreme near-collision subset.

\item
We present a unified risk annotation framework that combines physical risk modeling with LLM-based semantic reasoning, producing synchronized trajectory, metadata and natural-language representations for safety-critical driving.

\item We release a standardized data processing pipeline, benchmark protocols, and supporting tools for reproducible research on autonomous driving, including trajectory prediction, risk assessment, LLM-based reasoning, and decision-making.

\end{itemize}

\begin{table}[!ht]
\centering
\footnotesize
\setlength{\tabcolsep}{4pt}
\renewcommand{\arraystretch}{1.25}
\begin{tabular}{@{}p{1.9cm} p{1.8cm} p{1.5cm} p{2.0cm} p{3.1cm} p{1.7cm} p{2.0cm} p{0.55cm}@{}}
\toprule
\textbf{Dataset} & \textbf{Viewpoint} & \textbf{Modality} & \textbf{Source} & \textbf{Risk definition} & \textbf{Closed loop} & \textbf{Post-training stages} & \textbf{Year} \\
\midrule
\rowcolor{krours}
\textbf{K-Risk (ours)} & micro (BEV+ego) & trajectory \mbox{+text} & \mbox{EU+US+CN} (20 sources) & DRF top-10\%/1\% + $|a|$ thresholds + TTC$<$2/3/5s + 2s conflict predictor & yes (collision-free sim + reflection) & CPT + SFT + RLHF + RLVR & 2026 \\
\midrule
DriveLM~\cite{sima2024drivelm} & micro (ego) & image+graph & \mbox{US+SG+Sim} (2 sources) & none (safety implicit in P1--P3 QA) & no & SFT & 2024 \\
CoVLA~\cite{arai2024covla} & micro (ego) & video+text & JP (1 source) & none (free-text risk captions) & no & CPT + SFT & 2025 \\
OmniDrive~\cite{wang2025omnidrive} & micro (ego) & image+text & US+SG (1 source) & rule-based check on simulated trajectories & no & CPT + SFT & 2025 \\
Impromptu VLA~\cite{chi2025impromptuvla} & micro (ego) & video+text & \mbox{US+EU+CN+IN} (8 sources) & 4 unstructured-scenario categories (no quantitative score) & partial (NeuroNCAP scorer) & SFT + (partial RLVR) & 2025 \\
NuRisk~\cite{gao2025nurisk} & micro (BEV) & image+text & \mbox{US+SG+Sim} (3 sources) & weighted TTC + longitudinal/lateral DTC per agent & partial (CommonRoad sim) & SFT (CoT VQA) & 2025 \\
CODA-LM~\cite{li2024codalm} & micro (ego) & image+text & CN (1 source) & 7-class corner-case taxonomy (no kinematic threshold) & no & SFT & 2025 \\
InterHub~\cite{jiang2024interhub} & micro (BEV) & trajectory & \mbox{EU+US+CN} (4 sources) & MSAA $>0$ + PET annotation & no & none (event index only) & 2025 \\
LTD/UniVLT~\cite{huang2026ltd} & macro (roadside) & image+text & CN (1 source) & open-ended free-text hazard reasoning & no & SFT (curriculum) & 2026 \\
\bottomrule
\end{tabular}
\caption{Comparison of K-Risk with representative trajectory, VQA and VLA datasets for autonomous driving. Abbreviations: EU/US/CN/JP/SG/IN $=$ Europe/United States/China/Japan/Singapore/India; Sim $=$ simulator generated; CPT $=$ continued pre-training; SFT $=$ supervised fine-tuning; RLHF $=$ preference pairs for DPO/PPO/GRPO; RLVR $=$ reinforcement learning with verifiable simulator reward. Full column definitions are given in the surrounding text.}
\label{tab:dataset-comparison}
\end{table}%

\tabref{tab:dataset-comparison} compares K-Risk with representative trajectory, visual-question-answering (VQA), and vision-language-action (VLA) datasets. Two axes describe how K-Risk is positioned relative to these datasets. On the geographic axis, K-Risk combines real-world recordings from Europe, China and the United States, whereas most listed datasets are confined to a single region and the multi-region entries release event indices or question-answer pairs rather than verifiable safety signals. Aggregating 20 HV and AV trajectory sources across the three regions broadens the range of traffic rules, road geometries and driving conventions represented. On the risk and validation axes, K-Risk reports an explicit, multi-dimensional risk definition, built from a driver risk field, calibrated behavioral thresholds and a two-second trajectory-conflict predictor, together with closed-loop validation through a collision-free simulator and reflection.


\section*{Methods}
The K-Risk dataset was constructed from previously published trajectory data in two stages, described below. The Input data subsection lists the source datasets and their exact provenance. The annotation protocol is the set of rules and models that screens a continuous trajectory for high-risk events and assigns each retained event its semantic annotation. The data processing workflow is the end-to-end pipeline that applies this protocol to every source dataset and writes the per-event files. Throughout, the emphasis falls on how the records were produced and made reusable rather than on any single algorithmic component.

\subsection*{Input data}

K-Risk is a secondary dataset compiled from twenty publicly available trajectory sources, six recorded from human-driven vehicles (HV) and fourteen from automated-driving vehicles (AV). The six HV sources are highD~\cite{krajewski2018highd}, inD~\cite{bock2020ind} and rounD~\cite{bock2020round} from the LevelXData collection, recorded in Germany, the ExpresswayA and FreewayB subsets of CitySim~\cite{zheng2024citysim}, recorded in China and distributed through a collaboration documented in the CitySim release, and the I-80 recording of the NGSIM program~\cite{ngsim2006} in the United States. The fourteen AV sources are the Argoverse~2 Motion Forecasting dataset~\cite{wilson2023argoverse}, the Waymo Open Motion~\cite{ettinger2021womd} and Waymo Open Perception~\cite{sun2020waymo} datasets, the MicroSimACC dataset~\cite{yang2024microsimacc}, and the OpenACC Casale, Vicolungo, AstaZero and ZalaZONE recordings~\cite{makridis2021openacc}; the remaining AV sources, namely the CATS ACC, CATS Platoon and CATS UWM datasets, the Central Ohio single-vehicle and two-vehicle datasets, and the Vanderbilt ACC experiments, are obtained through the unified longitudinal trajectory collection of Ultra-AV~\cite{zhou2024ultraav}, which documents their original providers and access points. K-Risk releases only the event-level trajectory segments derived from these sources, not the complete original recordings, which remain available from their respective providers, listed with their repositories in the Data availability statement. Each source was used in accordance with its license and terms of use, and readers who require the full recordings obtain them from the original providers under the same terms. For each source the analysis uses the released vehicle trajectories at their native sampling rate and column schema, which the data processing workflow then converts into the common per-event representation.

\subsection*{Annotation protocol}

The annotation protocol screens a continuous trajectory for safety-critical events and then enriches every retained event with a semantic layer. Screening proceeds through three filters of increasing specificity, namely a driver risk field that scores the danger of each frame, a calibrated detector of behaviorally significant interactions, and a short-horizon trajectory-conflict predictor. The semantic layer that follows pairs each event with a structured scenario description and a set of abnormal-behavior notifications, and adds an LLM-generated risk analysis for a representative subset of the extreme events. The remainder of this subsection presents the two parts in turn.

The protocol first computes a driver risk field (DRF) value for each frame, which quantifies the danger experienced by the ego vehicle from three groups of factors. The first group is the ego-vehicle dynamics, comprising its speed, longitudinal and lateral acceleration, steering angle and heading. The second is the configuration of the surrounding vehicles, comprising the number of nearby agents, their classes such as car, truck, bus, pedestrian and cyclist, and their speed, acceleration and heading. The third is the spatial proximity to those vehicles, measured as Euclidean distances weighted by positional relevance across the preceding, following and adjacent lanes. Because road structure and traffic density vary across sources, the DRF threshold is set per source, and only the top 10\% of frames by DRF value are retained. This first filter removes routine conditions such as constant-speed highway cruising and concentrates the subsequent analysis on frames with higher interaction complexity and elevated potential risk; the full mathematical formulation of the DRF is given in \appref{appendix:drf_formulation}.

\begin{figure}[htbp]
    \centering
    \includegraphics[width=1\linewidth]{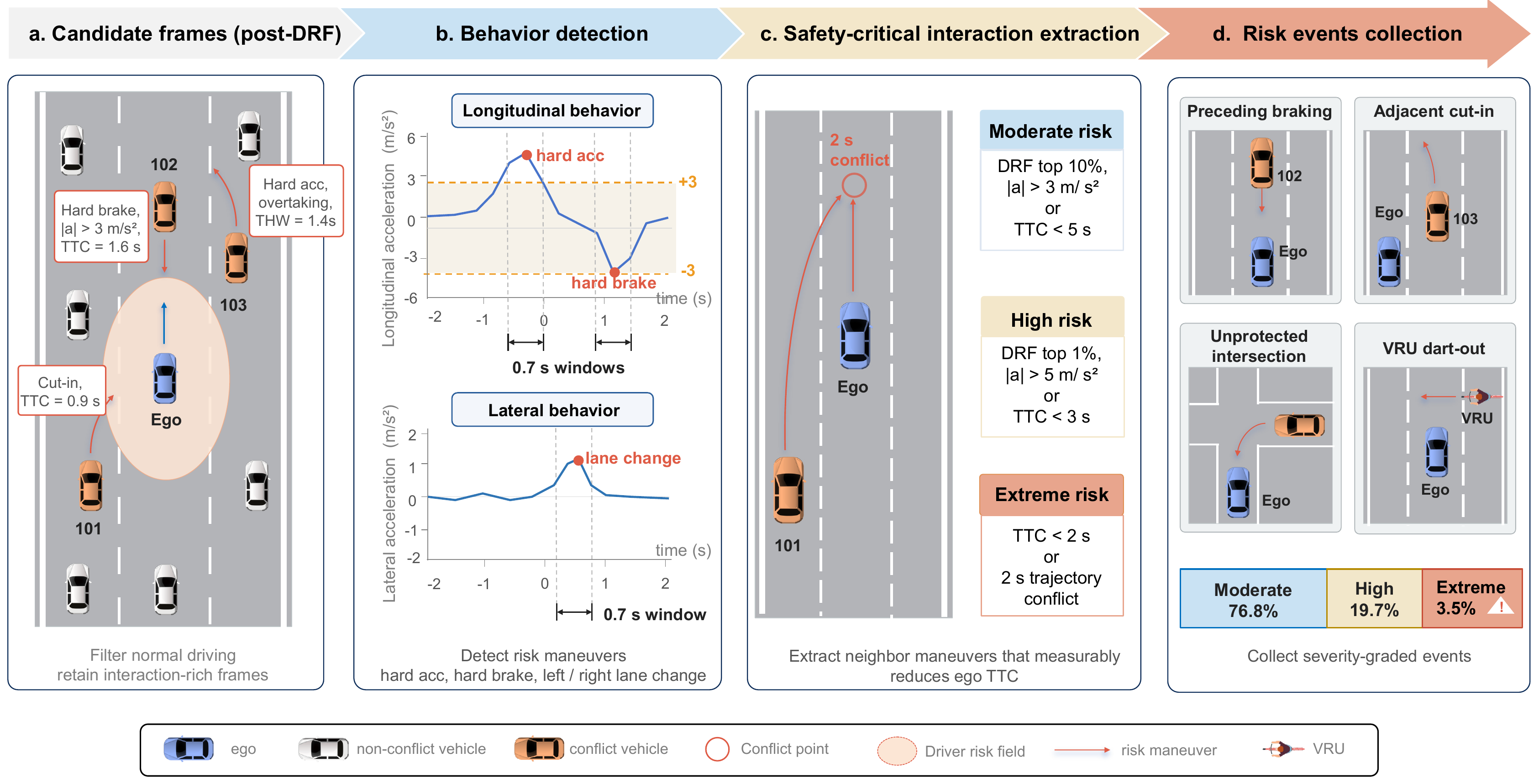}
    \caption{Detection of risk-relevant maneuvers and the resulting graded events. \textbf{(a)} The driver risk field retains interaction-rich candidate frames and removes routine cruising; the ego vehicle and its surrounding agents are shown with their TTC and time headway. \textbf{(b)} Each maneuver is detected within a $0.7$\,s window: longitudinal acceleration beyond $\pm 3\,\text{m/s}^2$ marks hard acceleration or hard braking, and a peak in lateral velocity marks a lane change. \textbf{(c)} An interaction is kept as safety-critical only when a neighboring vehicle measurably reduces the ego TTC, for example through a forecast trajectory conflict. \textbf{(d)} Retained events are graded as moderate (76.8\%), high (19.7\%) or extreme (3.5\%) risk by the calibrated thresholds, illustrated by preceding braking, adjacent cut-in, unprotected intersection and VRU dart-out cases.}
    \label{interaction}
\end{figure}

The second filter isolates interactions in which the surrounding vehicles exert a measurable influence on the ego vehicle, so that ego-initiated maneuvers unrelated to the environment are not retained. Four behavioral actions are treated as risk-relevant, namely hard acceleration, hard braking, left lane change and right lane change, and each is detected with a calibrated threshold, as illustrated in \figref{interaction}. Longitudinal acceleration above $+3\,\text{m/s}^2$ is labeled hard acceleration and acceleration below $-3\,\text{m/s}^2$ hard braking. These values, approximately $\pm 0.3g$, are widely used in traffic-safety research as indicators of abrupt or emergency driving and of passenger discomfort~\cite{ishikawa1991,liu2015traffic}. Lane changes are evaluated against dataset-specific lateral-speed profiles, because the severity of a lateral maneuver depends strongly on road geometry and longitudinal speed~\cite{milanes2010lanechange}. For each source the average lateral velocity during lane-change events is computed, and the threshold is set at 75\% of that value, which tolerates normal variability while filtering out gradual, non-critical lane shifts. Each maneuver is classified within a detection window of $0.7$ seconds, approximately the fastest documented human brake-reaction time, which keeps the detector responsive while avoiding intervals so short that sensor noise is mislabeled as behavior or so long that the immediacy of the action is lost~\cite{green2000reaction, treiber2013traffic}. An interaction is judged behaviorally significant only when such a maneuver produces a measurable escalation of risk consistent with traffic-rule interpretation. A left-preceding vehicle that changes into the ego lane and reduces the time-to-collision (TTC) below 5 seconds or the time headway below 2 seconds, for example, is flagged as critical, and more generally any behavior that reduces the TTC of the ego vehicle or of a surrounding vehicle in the current or an adjacent lane is considered significant, in line with established traffic-safety guidelines for situations that call for evasive or defensive maneuvers~\cite{ttc_timeheadway}.

The third filter complements TTC, which estimates the time to a potential collision under the current kinematic state but remains a one-dimensional measure. To capture the spatial dimension, the trajectories of the ego vehicle and the surrounding agents are forecast over a two-second horizon under constant-acceleration and constant-steering assumptions, a model well validated for short-term motion estimation~\cite{scholler2020constant, gilles2021review}. A conflict is declared when two forecast paths overlap within that horizon. This combination captures imminent longitudinal hazards together with the lateral conflicts that arise during lane changes, merges and weaving, and is especially useful in complex environments such as roundabouts and unsignalized intersections where several vehicles follow intersecting paths. Scenarios with TTC below 2~s or a detected trajectory conflict within two seconds are classified as extreme-risk events and placed in a dedicated subset for evaluating safety-critical decision-making.

The semantic layer then links the numerical records of each retained event to interpretable text, so that the event is usable not only for conventional trajectory modeling but also for language-based risk interpretation and decision reasoning. For every event the protocol generates two rule-based records, namely a structured scenario description and a set of abnormal-behavior notifications, and for a representative subset of the extreme events it adds a third record, an LLM-generated risk analysis. Together these records describe the road context, the traffic agents involved, the risk-relevant behaviors and the recommended response, in a form that a driving agent can parse directly or that can serve as a supervised language-model training sample. The structured scenario description supplies the contextual information required to interpret the event. It records the road layout, lane topology, speed limits and the legal maneuvers available to the ego vehicle. It then describes the ego vehicle together with up to eight surrounding agents, namely the preceding, left-preceding, right-preceding, left-alongside, right-alongside, following, left-following and right-following agents, giving for each its relative position, lane relation, velocity, acceleration, heading and vehicle class where available. This representation mirrors the way a human driver observes surrounding traffic while preserving machine-readable state variables for model training and evaluation.

The abnormal-behavior notifications are derived from the event-extraction results and attach concise risk cues to the description. Each agent receives behavior labels when it triggers a predefined risk-relevant pattern such as hard braking, hard acceleration, an abrupt lane change, a short time-to-collision or a trajectory overlap. When such behavior produces a meaningful interaction with the ego vehicle, for example a cut-in from a left-preceding vehicle that reduces the time-to-collision below a safety threshold, the corresponding warning is attached to the event record. Isolated abnormal behaviors such as excessive speed or close-proximity following are recorded even when no explicit interaction is detected. Additional cautionary messages are added when vulnerable road users or large vehicles, namely pedestrians, cyclists, trucks, buses or motorcycles, appear in the surrounding context, because such agents impose stricter safety margins, larger blind zones or different right-of-way considerations. These notifications reduce the chance that a language model overlooks important but sparse safety signals, and they let the dataset represent mixed-traffic interactions in a way compatible with both traffic-safety analysis and language-based scenario understanding.

For the representative subset, an LLM generates an event-level risk analysis from the structured description and the abnormal-behavior notifications. The model receives the lead-in phase, the peak-risk frame and the following phase, together with the surrounding-agent states and the rule-based risk reminders. The generated text explains what led to the hazardous state, why the event is safety-critical, how the interaction evolves over time and which response is recommended. The response is constrained by a predefined action schema, namely keeping the current state, changing lane to the left or right, accelerating or decelerating. This schema links the textual reasoning to a discrete decision label that can be evaluated in downstream experiments. To improve reliability, the LLM-generated records were reviewed by domain experts for factual correctness, contextual consistency and action validity, and the feedback was used to refine the prompt template, adjust the risk reminders and correct ambiguous descriptions. The same generation procedure is released with the dataset, so that the LLM layer can be reproduced and extended to additional events. The final semantic annotation therefore combines rule-based risk cues, structured traffic context and natural language reasoning, producing event records that support risk interpretation, behavior modeling, decision prediction and language-based reasoning.

\subsection*{Data processing workflow}

\begin{figure}[h]
    \centering
    \includegraphics[width=1\linewidth]{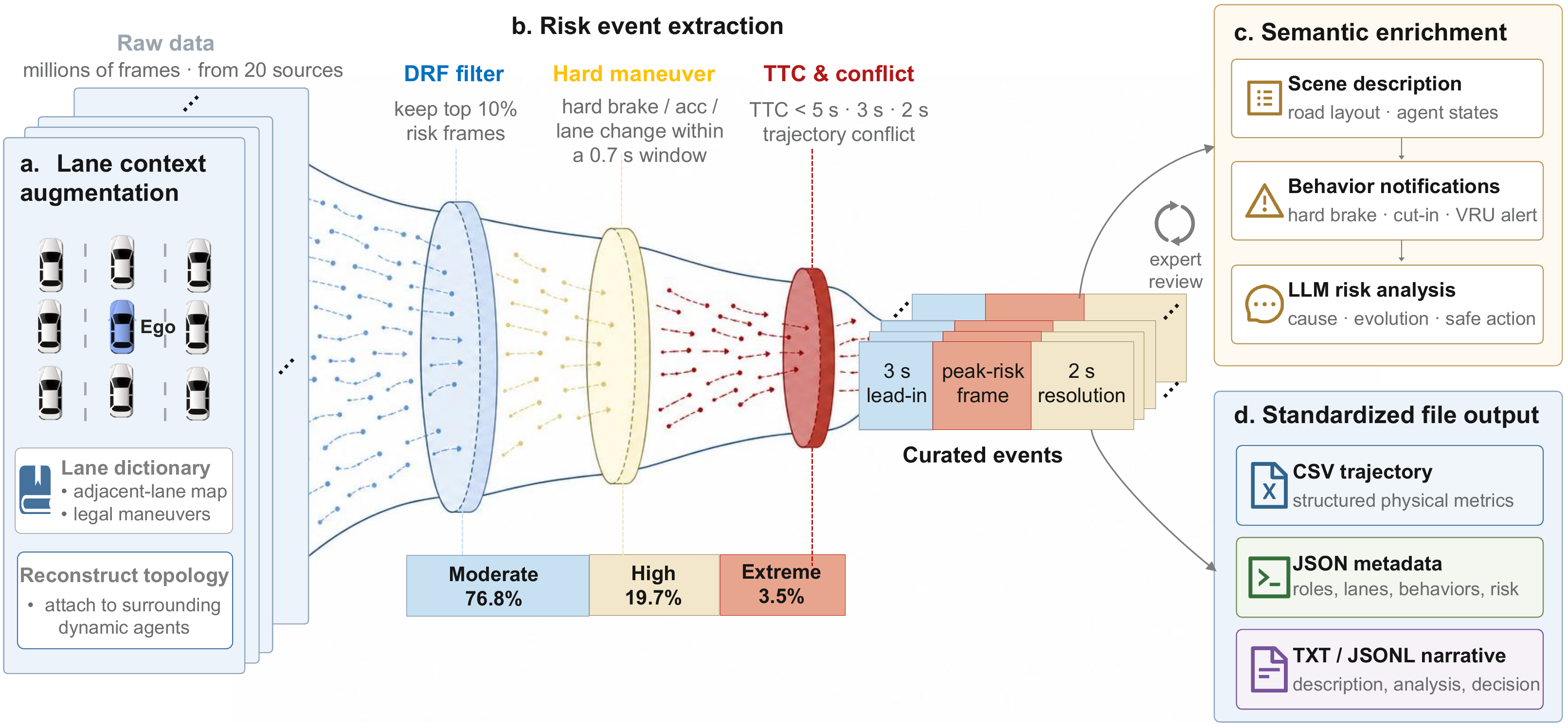}
    \caption{Data processing workflow of K-Risk. Raw records from the 20 sources first pass through lane context augmentation, which reconstructs the lane topology and attaches up to eight surrounding agents to each ego vehicle (a). Risk event extraction then applies the driver risk field filter, the hard-maneuver detector and the TTC and trajectory-conflict test, and grades the retained events as moderate (76.8\%), high (19.7\%) or extreme (3.5\%) risk, with each event window comprising a $3$\,s lead-in, a peak-risk frame and a $2$\,s resolution phase (b). Each curated event is enriched with a scenario description and behavior notifications, and a representative subset additionally receives an LLM risk analysis with expert review in the loop (c), and written to a standardized three-part file: a CSV of trajectories, a JSON of metadata and a text narrative (d).}
    \label{fig:processing_flow}
\end{figure}

The data processing workflow applies the annotation protocol to each source dataset and writes the resulting per-event files, proceeding through lane-topology construction, surrounding-vehicle augmentation, event extraction with risk grading, and annotation enrichment, as shown in \figref{fig:processing_flow}. The same workflow runs on every source, so that the heterogeneous upstream recordings yield events in a single, consistent format.

On loading the raw trajectory data from a source, the workflow first builds a structured dictionary that encodes the lane topology. Most datasets provide lane identifiers but not the positional relationships between lanes, such as adjacent-lane mappings, so this step reconstructs the spatial configuration of the road network and makes it possible to locate vehicles relative to the ego vehicle. Using this dictionary, the workflow appends up to eight surrounding-vehicle identifiers to each ego-vehicle instance, covering the preceding, following, left- and right-preceding, left- and right-following, and left- and right-alongside positions. This representation mirrors the situational awareness of a driver and provides the role structure on which the semantic annotation depends.

The workflow then extracts events and grades their severity. It applies the DRF filter to discard routine, uneventful segments and detects hard maneuvers over the $0.7$-second window using the calibrated thresholds of the annotation protocol. A frame is flagged as risk-relevant only when it shows both a significant behavioral indicator, such as hard acceleration, braking or an abrupt lane change, and a temporal-proximity measure below a safety threshold, such as a low TTC or a projected trajectory conflict. Each risk-relevant frame then receives a severity grade. A frame is graded high risk when its DRF lies within the top 1\% of the source distribution, the magnitude of its acceleration or braking exceeds $5\,\text{m/s}^2$, or its TTC falls below 3~s, and it is graded moderate risk when the DRF lies within the top 10\%, the acceleration or braking magnitude exceeds $3\,\text{m/s}^2$, or the TTC falls below 5~s; extreme-risk frames, identified by a TTC below 2~s or a trajectory conflict within two seconds, are graded separately as defined in the annotation protocol. Around each graded frame the workflow extracts an event window comprising the 3 seconds preceding and the 2 seconds following it, which captures the full temporal context in three phases: a lead-in phase that exposes anticipatory behavior before the risk peak, a peak-risk frame that represents the highest-risk state, and a resolution phase that records the actual response of the human driver or autonomous system. This segmentation lets downstream models learn both risk-escalation patterns and the causal relationship between decisions and their outcomes.

For each extracted event the workflow finally appends the multi-layer annotation defined by the protocol and writes the event to a standardized three-part file schema. The annotation comprises the structured scenario description and the abnormal-behavior notifications for every event, together with the LLM-generated risk analysis for the representative subset described above. The CSV file holds the per-event trajectory records, including position, velocity, heading and acceleration for every relevant agent; the JSON file encodes the symbolic event metadata, including agent roles, lane relationships, scenario context, interaction structure, behavioral labels and the assigned severity grade; and the text file, in JSON Lines format, holds the natural language narrative, comprising the environment description and, where available, the risk analysis and the recommended decision. This layered structure supports conventional supervised learning for motion prediction and trajectory forecasting as well as language-based reasoning and decision-making, and the precise field definitions of each file are listed in \appref{appendix:schema}.

\section*{Data Records}

Each data record in the K-Risk dataset is a multi-modal representation that combines per-event trajectory segments, structured metadata and natural language semantic descriptions, with visualization snapshots provided for inspection. This design supports analysis across diverse tasks, including behavior modeling, risk-sensitive decision-making and LLM-based prompt engineering.

\subsection*{Event Structure}

\begin{figure}[h]
\centering
\includegraphics[width=1\linewidth]{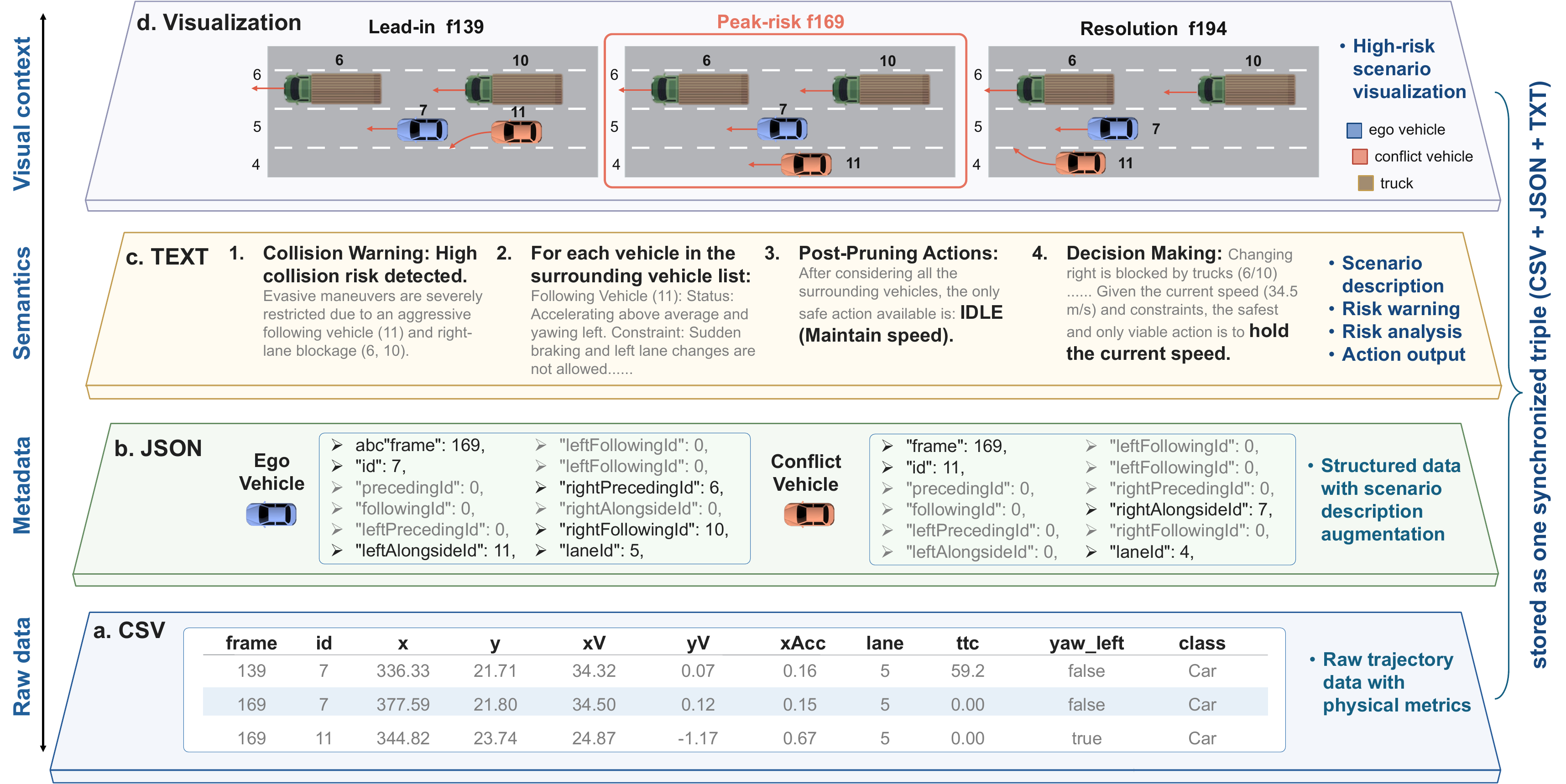}
\caption{Layered structure of a single K-Risk event, stored as one synchronized triple. The CSV layer holds the raw per-frame trajectories of the ego and conflict vehicles, including position, velocity, acceleration, lane, TTC and behavioral flags (a). The JSON layer encodes the structured metadata, with agent roles, lane relationships and the risk level (b). The text layer provides the natural language annotation, comprising the imminent-collision warning, the surrounding-agent analysis, the post-pruning actions and the final decision (c). Visualization snapshots of the lead-in, peak-risk and resolution phases support inspection (d).}
\label{fig:event_structure}
\end{figure}

As shown in~\figref{fig:event_structure}, every event is extracted from a continuous trajectory and segmented into a lead-in phase, a peak-risk frame and a resolution phase. The peak-risk frame corresponds to the highest-risk point, such as the moment of minimum time-to-collision, while the lead-in and resolution phases provide the temporal context necessary for interaction analysis.

Each event is stored in three synchronized formats to ensure compatibility across research tasks. The CSV file contains the per-event trajectory records for all participating agents, including their spatial coordinates, velocity components, acceleration values, lane identifiers and heading. The JSON file encodes the structured metadata, capturing lane topology, agent roles, detected behaviors, time-to-collision estimates and the assigned risk level. The text file provides the semantic annotation in natural language, comprising the environment description and, for the representative subset, the detailed risk analysis and the recommended decision. The complete field definitions of each file are listed in \appref{appendix:schema}.

\subsection*{File organization and access}

K-Risk is hosted in the public repository described in the Data availability statement. The release is organized into three top-level directories. The \filepath{trajectory_data} directory holds the per-event trajectory segments that underlie each retained event, the \filepath{event_annotations} directory holds the extracted per-event records, and the \filepath{llm_analysis} directory holds the natural language scenario descriptions and the corresponding LLM responses. Within \filepath{trajectory_data} and \filepath{event_annotations} the events are separated into human-driven (\filepath{HV}) and automated-driving (\filepath{AV}) groups, then into one folder per source dataset. The human-driven sources are further split into risk-level folders, for example \filepath{highd_normal_risk} and \filepath{highd_high_risk}, and the near-collision cases are collected in a separate \filepath{extreme} folder. Each per-event triple shares the event identifier encoded in its file name, so that the CSV, JSON and text records of one event can be matched across directories. The complete original recordings are not redistributed; they remain available from their providers and are cited as input sources in the Methods Input data subsection and listed in the Data availability statement~\cite{krajewski2018highd,bock2020ind,bock2020round,zheng2024citysim,ngsim2006,wilson2023argoverse,ettinger2021womd,sun2020waymo,yang2024microsimacc,makridis2021openacc,zhou2024ultraav}.


\section*{Data Overview}

\begin{figure}[!t]
\centering
\includegraphics[width=1\linewidth]{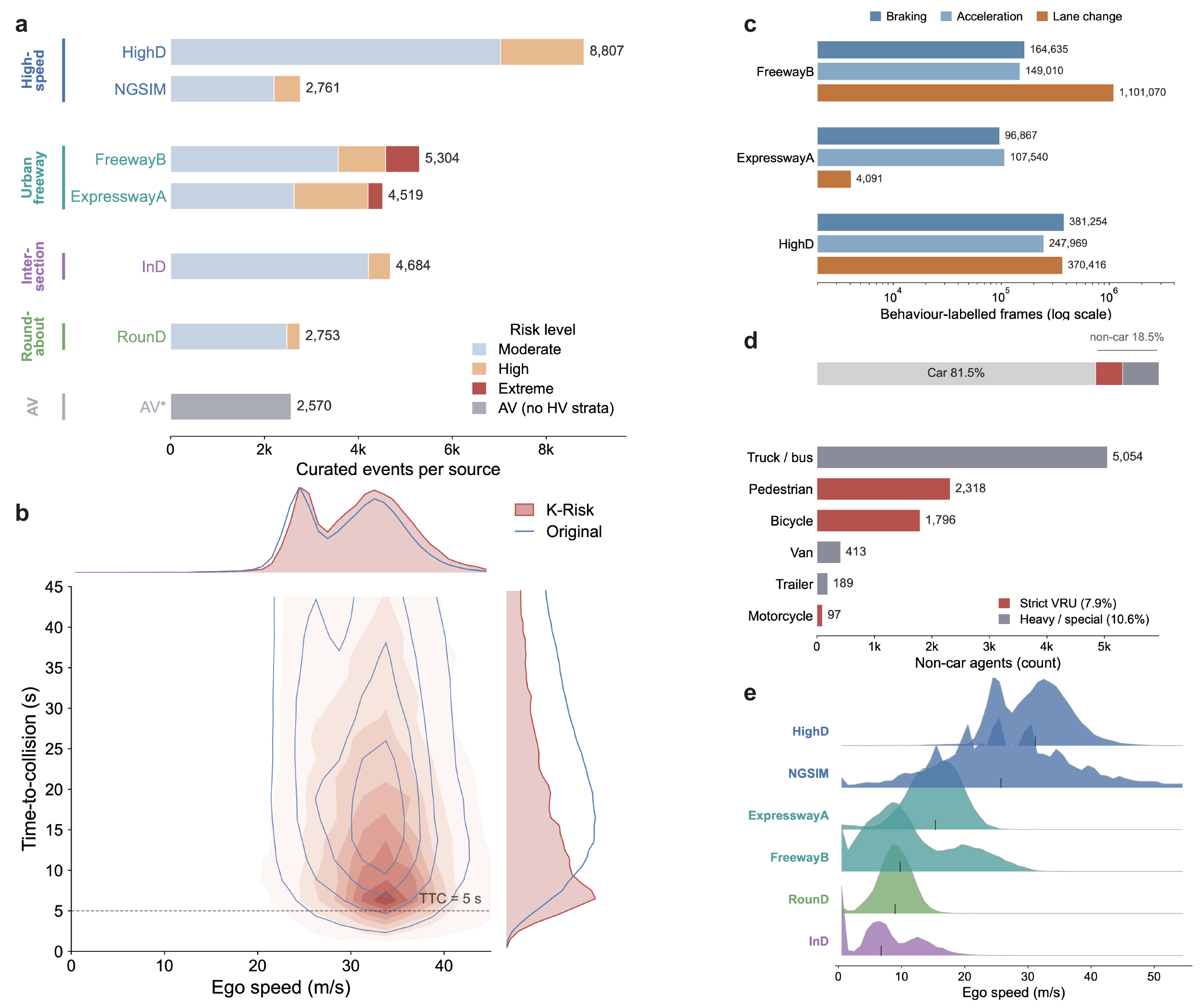}
\caption{\textbf{Composition and safety-critical properties of K-Risk.}
\textbf{(a)} Curated events per source, stacked by risk level (moderate, high, extreme) and grouped by driving environment; the autonomous-vehicle (AV) subset carries no human-driver risk stratification.
\textbf{(b)} Joint ego-speed $\times$ time-to-collision (TTC) density for K-Risk (filled) versus the original HighD data (blue contours), with speed and TTC marginals; K-Risk concentrates below the $\mathrm{TTC}=5$\,s line (dashed).
\textbf{(c)} Behaviour-labelled frame counts per source (log scale) for braking, acceleration and lane change.
\textbf{(d)} Agent composition over $53{,}295$ unique agents (top), with the non-car classes resolved by count (bottom).
\textbf{(e)} Per-scenario ego-speed distributions (ridgeline, peak-normalised, ordered by median; the tick marks the median).}
\label{fig:dataset_distribution}
\label{fig:data_distribution}
\label{fig:ttc_speed_distribution}
\end{figure}

The complete K-Risk dataset contains 31,398 curated events, systematically categorized by their source datasets and assigned risk levels, with the distributions shown in \figref{fig:dataset_distribution}a. Although the events are drawn from the 20 upstream trajectory sources, the high-risk yield is uneven: a small number of large naturalistic recordings dominate, because high-risk events are sparse in many sources and especially rare in the autonomous-vehicle recordings. The bulk of the events therefore originate from a handful of sources, with highD contributing the largest share at 8,807 events, followed by FreewayB at 5,304, inD at 4,684 and ExpresswayA at 4,519; NGSIM, the roundabout recordings and the aggregated autonomous-vehicle sources add a further 2,761, 2,753 and 2,570 events, respectively. Each human-driven event is assigned one of three risk levels, moderate (76.8\%), high (19.7\%) and extreme (3.5\%), with the percentages taken over the mutually exclusive human-driven events, and the extreme fraction corresponds to the 1,036-event extreme subset, while the autonomous-vehicle events are released without a human-driver risk stratification. These classifications follow the calibrated thresholds of the annotation protocol, which combine time-to-collision metrics, driver risk field percentiles and the detection of hard maneuvers such as abrupt braking or lane changes. \figref{fig:dataset_distribution} further reports the agent composition and the per-scenario speed and time-to-collision distributions, which together summarize the heterogeneous traffic participants and the range of severity levels present in the release.

\section*{Technical Validation}

We validate the K-Risk dataset along four dimensions: the diversity of its scenarios and participants, its emphasis on high-risk edge cases, its compatibility with LLM-based agents for decision-making, and the experimental performance of such an agent on the extreme subset.

\subsubsection*{Diversity of scenarios and participants}

To reflect real-world complexity, K-Risk spans diverse road structures, participant types, and risk-relevant behaviors. Across the human-driven events, the driving environments include high-speed highways (40.1\%), urban freeways (34.1\%), intersections (16.2\%), and roundabouts (9.5\%), as shown in \figref{fig:data_distribution}a. Behavioral variability is also broad. In the three sources that provide frame-level behavior labels, braking, acceleration, and lane changes are each marked on $10^5$ to $10^6$ frames, as shown on the logarithmic scale of \figref{fig:data_distribution}c. Unlike datasets focused mainly on passenger vehicles, K-Risk includes vulnerable road users (VRUs) and heavy-duty vehicles. Among 53,295 unique agents, 18.5\% are non-car road users (\figref{fig:data_distribution}d): VRUs, including pedestrians, cyclists, and motorcyclists, account for 7.9\%, while heavy or special vehicles, including trucks, buses, vans, and trailers, account for a further 10.6\%. This diversity allows K-Risk to support interaction-rich, mixed-traffic, and safety-critical evaluations of autonomous systems.

\subsubsection*{Emphasis on high-risk edge cases}
K-Risk prioritizes edge scenarios where collision likelihood is elevated. Compared with the original datasets, K-Risk exhibits markedly lower time-to-collision (TTC) values, obtained through calibrated filtering thresholds (for example, TTC $<$ 5 s and top-percentile DRF scores). Because these thresholds enter the extraction protocol, the shift toward low TTC is expected by construction; the joint speed and TTC density in \figref{fig:ttc_speed_distribution}b illustrates its magnitude, showing that relative to the original HighD records the K-Risk events concentrate in the near-collision region below the $\mathrm{TTC}=5$\,s line while retaining a realistic speed spread rather than collapsing onto the threshold, which supports the evaluation of time-critical decision-making algorithms. The dataset also preserves coverage across all speed regimes. As the per-scenario speed distributions in \figref{fig:ttc_speed_distribution}e show, urban intersection and roundabout scenarios concentrate at low speed, whereas the highway and freeway scenarios extend beyond 40 m/s, supporting analysis of high-speed merges, lane changes, and emergency maneuvers. This range allows planning and control systems to be tested under both congested and high-speed conditions.

\subsubsection*{Compatibility with LLM-based agents}
\begin{figure}[h]
    \centering
    \includegraphics[width=1\linewidth]{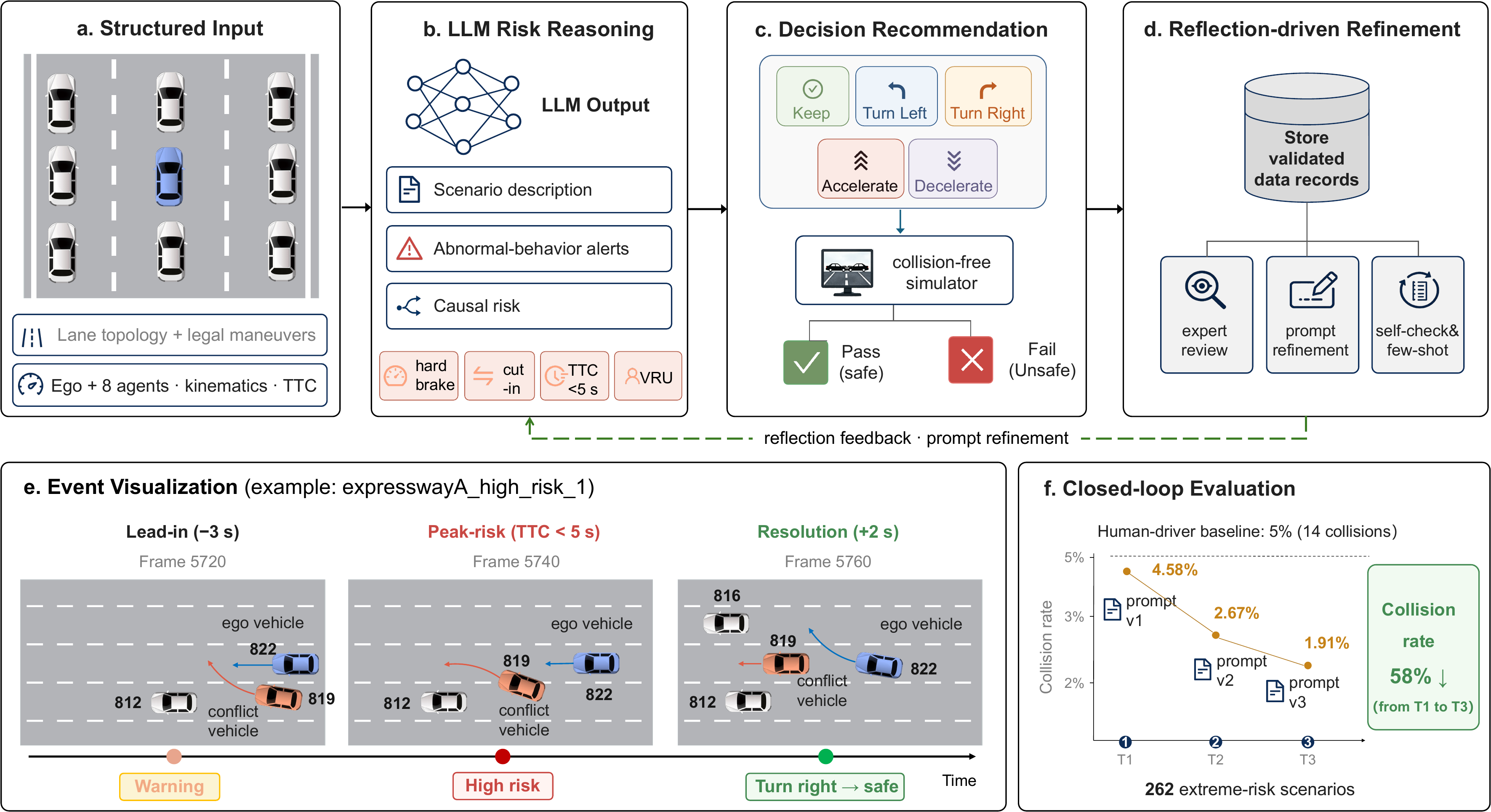}
    \caption{Closed-loop LLM annotation framework. A structured input describes the ego vehicle, up to eight surrounding agents, the lane topology and legal maneuvers, and the TTC (a). The LLM agent reasons over the scenario description, the abnormal-behavior notifications and the causal risk cues (b) and recommends an action from the five-action schema (c). The action is checked against a collision-free simulator, and a failure triggers reflection that refines the prompt and yields validated records through expert review and self-check (d). A real ExpresswayA event is shown across its lead-in, peak-risk and resolution phases (e), and the collision rate on the 262 extreme-risk evaluation events is $4.58\%$, $2.67\%$ and $1.91\%$ over three trials, shown together with the $5\%$ human-driver reference (f). This event is the worked example detailed in the stages below.}
    \label{fig:llm_annotation_loop}
\end{figure}

K-Risk is accompanied by a closed-loop LLM annotation framework for scalable and interpretable decision reasoning, illustrated in~\figref{fig:llm_annotation_loop}. In this framework a high-risk scenario is converted into a structured description of the ego and surrounding vehicles, the legal maneuvers, and the interaction context. The LLM generates risk-aware decisions, such as ``Turn Right'' or ``Decelerate,'' with justifications based on spatial relations and predicted hazards. All actions are validated through collision-free simulation, and the feedback is fed back to refine the prompts. Validated outcomes form a reusable memory pool that supports few-shot learning for LLM-based driving agents. This design makes the dataset compatible with closed-loop experiments and allows decision-making to be evaluated under safety-critical conditions. A complete walk-through of one event from this framework, covering the system prompt, the structured scenario description fed to the LLM, the rule-based risk reminders produced by the risk filters, and the LLM-generated risk analysis together with the chosen action identifier, is provided in \appref{appendix:closed_loop_walkthrough}.

\subsubsection*{Closed-loop usability of the extreme subset}

The extreme subset was further evaluated to examine whether K-Risk can support closed-loop, safety-critical decision testing. We selected 262 extreme-risk scenarios that include severe near-collision interactions and 14 recorded human-driver collisions. For each scenario, an LLM-based decision-making agent received the structured scene description, abnormal-behavior notifications, and causal risk cues, and then selected one action from the five-action schema. The selected action was evaluated by a collision-free simulator. Failed cases were returned to the agent as reflection feedback, and the refined prompt was used in the next trial.
The collision rate decreased from 4.58\% in the first trial to 2.67\% in the second trial and 1.91\% in the third trial, corresponding to a 58.3\% relative reduction from Trial 1 to Trial 3 (\tabref{tab:closed-loop-validation}). The non-zero collision rate across all trials indicates that the subset is not composed of routine driving cases, but instead contains scenarios that remain challenging even after iterative refinement. More importantly, each tested event produces reusable validation records, including the initial decision, simulator outcome, failure reason, reflection prompt, and refined decision. These records allow K-Risk to serve not only as a collection of high-risk trajectories, but also as a stress-test resource for evaluating risk-aware decision-making, prompt refinement, and preference-based learning.

\begin{table*}[!ht]
\centering
\normalsize
\setlength{\tabcolsep}{6pt}
\renewcommand{\arraystretch}{1.15}
\begin{tabular}{@{}lcccc@{}}
\toprule
\textbf{Trial} & \textbf{Evaluated scenarios} & \textbf{Collisions} & \textbf{Safe outcomes} & \textbf{Collision rate (\%)} \\
\midrule
Human driver & 262 & 14 & 248 & 5.34 \\
Trial 1 & 262 & 12 & 250 & 4.58 \\
Trial 2 & 262 & 7 & 255 & 2.67 ($-$41.7\%) \\
Trial 3 & 262 & \textbf{5} & \textbf{257} & \textbf{1.91 ($-$58.3\%)} \\
\bottomrule
\end{tabular}
\caption{Closed-loop validation on the K-Risk extreme subset.}
\label{tab:closed-loop-validation}
\end{table*}

\begin{table}[!ht]
\centering
\footnotesize
\setlength{\tabcolsep}{4pt}
\renewcommand{\arraystretch}{1.25}
\begin{tabularx}{\linewidth}{@{}p{2.4cm} p{4.4cm} p{5cm} p{5cm}@{}}
\toprule
\textbf{Family} & \textbf{Task} & \textbf{K-Risk asset used} & \textbf{Suggested benchmark / scorer} \\
\midrule
\multirow{3}{2.4cm}{\textbf{Perception \& reasoning}}
 & Risk-level classification (Moderate/High/Extreme) & JSON severity grade & internal macro-F1; CODA-LM~\cite{li2024codalm} \\
 & Risk-cue VQA (which agent, why) & TXT risk explanations $\rightarrow$ QA pairs & Lingo-Judge~\cite{marcu2023lingoqa}; DriveLM~\cite{sima2024drivelm}; DVBench~\cite{zeng2025dvbench} \\
 & Causal / counterfactual reasoning & three-phase lead-in/critical/following slice & OmniDrive~\cite{wang2025omnidrive} counterfactual QA \\
\midrule
\multirow{3}{2.4cm}{\textbf{Prediction}}
 & Intent prediction & JSON behavior labels & DRAMA-X~\cite{godbole2025dramax} \\
 & Trajectory forecasting & CSV trajectory & ADE/FDE on Argoverse~2, WOMD \\
 & Risk / TTC forecasting & frame-level TTC in CSV/JSON & InterHub PET metric~\cite{jiang2024interhub} \\
\midrule
\multirow{4}{2.4cm}{\textbf{Decision \& action}}
 & Discrete action prediction (5-class) & TXT decision recommendation & internal accuracy / macro-F1 \\
 & Closed-loop driving & 1{,}036-event extreme subset as stress test & NeuroNCAP~\cite{ljungbergh2024neuroncap}; Bench2ADVLM~\cite{zhang2025bench2advlm} \\
 & Plan justification & TXT risk analyses as references & BLEU / ROUGE / CIDEr; Lingo-Judge~\cite{marcu2023lingoqa} \\
 & Preference / reflection learning & trial-1 vs trial-3 chosen/rejected pairs & DPO~\cite{rafailov2023dpo} / SimPO~\cite{meng2024simpo} training; pairwise win-rate \\
\midrule
\multirow{3}{2.4cm}{\textbf{Evaluation \& audit}}
 & Interpretable safety evaluation & 1{,}036 events extreme subset & collision rate, infraction count \\
 & Responsibility analysis & JSON agent roles + TXT causal chain & expert inter-rater agreement \\
 & Rule-compliance audit & JSON legal-action schema vs LLM output & rule-violation rate \\
\midrule
\multirow{2}{2.4cm}{\textbf{Data augmentation}}
 & Scenario augmentation (counterfactual variants) & CSV trajectory + TXT description as seed & CrashAgent~\cite{wang2025crashagent}; ChatScene~\cite{zhang2024chatscene} \\
 & Curriculum construction for RL & three risk levels as difficulty stages & internal RL curriculum \\
\bottomrule
\end{tabularx}
\caption{Downstream tasks that K-Risk supports, grouped into five families. The ``K-Risk asset'' column indicates which of the per-event CSV, JSON and text files, or the extreme subset and trial pairs, provides supervision; the ``Suggested benchmark/scorer'' column lists external evaluators or established metrics that are compatible with the K-Risk release.}
\label{tab:downstream}
\end{table}

\section*{Usage Notes}

K-Risk is released as a set of per-event triples together with the extreme subset, and the file schema in \appref{appendix:schema} is intended to let others reuse the data without re-running the construction pipeline. The three files per event support complementary modes of use: the CSV trajectories feed conventional supervised learning for motion prediction and trajectory forecasting, while the JSON metadata and natural language text support language-based risk interpretation and decision-making. \tabref{tab:downstream} summarizes the downstream tasks that the release supports, grouped into five families, and indicates which asset each task consumes together with a suggested external benchmark or scorer where one exists. Each suggested evaluator corresponds to a published benchmark or metric that can be applied to the K-Risk release. These evaluators include NeuroNCAP~\cite{ljungbergh2024neuroncap} for closed-loop driving, CODA-LM~\cite{li2024codalm} for corner-case classification, Lingo-Judge~\cite{marcu2023lingoqa} and DVBench~\cite{zeng2025dvbench} for risk-cue question answering, DriveLM~\cite{sima2024drivelm} for graph-structured perception, prediction and planning, DRAMA-X~\cite{godbole2025dramax} for fine-grained intent prediction, OmniDrive~\cite{wang2025omnidrive} for counterfactual reasoning, and the post-encroachment-time metric of InterHub~\cite{jiang2024interhub} for risk forecasting.

Beyond these analysis tasks, the structure of K-Risk aligns with the standard
stages used to adapt a general-purpose LLM into a driving agent
~\cite{ouyang2022instructgpt,rafailov2023dpo,meng2024simpo,lambert2024rlvr}.
The structured natural language descriptions can serve as a continued
pre-training corpus that exposes a base model to driving-specific vocabulary,
road-rule context, interaction patterns and quantitative agent states. For the
subset that carries an LLM annotation, the description, risk analysis and
decision jointly form a supervised fine-tuning sample that links the
five-action schema to a causal explanation of the driving risk. Closed-loop
validation further produces trial-level outputs, with the collision rate
falling from 4.58\% in the first trial to 1.91\% in the third trial on the
extreme cases. These first-trial and third-trial outputs naturally form
rejected and chosen response pairs that can support preference optimization,
such as Direct Preference Optimization~\cite{rafailov2023dpo} or Simple
Preference Optimization~\cite{meng2024simpo}, without additional human
labeling. The collision-free simulator used to validate each annotation can
also provide a verifiable reward signal for reinforcement learning
~\cite{lambert2024rlvr,shao2024deepseekmath}, while the explicit five-action
schema enables constrained decoding at inference time. In this way, K-Risk can
support a complete adaptation pipeline from domain exposure and supervised
alignment to preference learning and reward-based refinement for risk-aware
driving agents.

Two properties of the release should be kept in mind when reusing the data. High-risk events involving autonomous vehicles are comparatively rare, because the automated-driving sources exhibit smoother driving behavior, so the autonomous-vehicle component is smaller than the human-driven component. The risk definition rests on calibrated thresholds over physical and behavioral indicators rather than on a single universal standard, and it does not encode probabilistic uncertainty or multi-agent intent.


\section*{Data availability}

The K-Risk dataset is deposited in a public repository and is available at \url{https://github.com/benmagnifico/K-Risk} (code and processed records) and at the Figshare record \url{https://doi.org/10.6084/m9.figshare.32896772}. The release contains the event-level trajectory segments, the risk annotations and metadata indices, the natural language semantic descriptions and LLM responses, and the benchmark splits, organized into the \filepath{trajectory_data}, \filepath{event_annotations} and \filepath{llm_analysis} directories described in the Data Records section.

K-Risk is a secondary dataset compiled from publicly available human-driven (HV) and automated-driving (AV) trajectory sources, which are cited here as input data and remain available from their original providers; they are not redistributed as part of K-Risk. The six HV sources are highD~\cite{krajewski2018highd} (\url{https://www.highd-dataset.com}), inD~\cite{bock2020ind} (\url{https://www.ind-dataset.com}) and rounD~\cite{bock2020round} (\url{https://www.round-dataset.com}) from the LevelXData collection, the ExpresswayA and FreewayB subsets of CitySim~\cite{zheng2024citysim} (\url{https://github.com/UCF-SST-Lab/UCF-SST-CitySim1-Dataset}), and the I-80 recording of NGSIM~\cite{ngsim2006} (\url{https://ops.fhwa.dot.gov/trafficanalysistools/ngsim.htm}). The fourteen AV sources are the Argoverse~2 Motion Forecasting dataset~\cite{wilson2023argoverse} (\url{https://www.argoverse.org/av2.html}), the Waymo Open Motion~\cite{ettinger2021womd} and Waymo Open Perception~\cite{sun2020waymo} datasets (\url{https://waymo.com/open/}), the MicroSimACC dataset~\cite{yang2024microsimacc} (\url{https://github.com/microSIM-ACC/ICE}), the OpenACC Casale, Vicolungo, AstaZero and ZalaZONE recordings~\cite{makridis2021openacc} (\url{https://data.jrc.ec.europa.eu/dataset/9702c950-c80f-4d2f-982f-44d06ea0009f}), and the CATS ACC, CATS Platoon and CATS UWM datasets, the Central Ohio single-vehicle and two-vehicle datasets, and the Vanderbilt ACC experiments, obtained through the unified Ultra-AV collection~\cite{zhou2024ultraav} (\url{https://github.com/CATS-Lab/Filed-Experiment-Data-ULTra-AV}).

\section*{Code availability}

The data processing pipeline that builds K-Risk is openly available in the same repository (\url{https://github.com/benmagnifico/K-Risk}). The repository documents the steps used to convert each source recording into the per-event triple and to reproduce the figures and statistics reported here. The event-extraction components implement the annotation protocol of the Methods, namely the driver risk field computation with its per-source top-percentile thresholds, the hard-maneuver detector over the $0.7$-second window, and the time-to-collision and two-second trajectory-conflict test that assign the severity grades. The source-specific scenario generators (\filepath{generate_highd.py}, \filepath{generate_expresswayA.py}, \filepath{generate_freewayB.py}, \filepath{generate_ind.py}, \filepath{generate_round.py}, \filepath{generate_ngsim.py} and \filepath{generate_AV.py}) build the lane context, attach the surrounding agents and write the structured scenario descriptions; \filepath{prompts.py} holds the system prompts and the five-action schema; \filepath{llm_test.ipynb} sends each scenario description to the LLM and stores the generated risk analysis and action; the closed-loop validation component forwards each recommended action to the collision-free simulator and records the per-trial outcomes and the reflection prompts, from which the first-trial and third-trial pairs are written; and \filepath{compute_statistics.py} and \filepath{plot_overview.py} compute the per-source and risk-level statistics and produce the composition figure. The code is written in Python~3 and relies on the standard scientific stack (\texttt{pandas}, \texttt{numpy} and \texttt{matplotlib}) together with an LLM client for the annotation step. The code is released for research use, and access to it carries no further restriction.

\bibliography{sample}


\section*{Acknowledgements}

This research was conducted by KAIST as part of a joint research project under the Korea Institute of Science and Technology Information (KISTI) R\&D program, ``Development of the Next-Generation Integrated Wired/Wireless Communication Gateway (X-Gateway).''

\section*{Author contributions}
H.H. conceived and supervised the project. H.H., J.L., and Z.Z. designed the K-Risk dataset and its construction protocol. J.L. extracted high-risk events and organized the dataset. Z.Z. curated the semantic annotations and implemented the data processing pipeline. M.W. conducted data preprocessing, and visualization. P.L., K.J., and J.W. provided technical guidance and revised the manuscript. All authors contributed to the manuscript and approved the final version.

\section*{Competing interests}

The authors declare no competing interests.



\newpage
\appendix

\section{Mathematical Formulation of the Driver Risk Field (DRF)}
\label{appendix:drf_formulation}

The DRF assigns to every frame a scalar that quantifies the danger the ego vehicle perceives from its own motion and from the surrounding agents. K-Risk adopts the speed- and steering-adaptive DRF of SafeDrive~\cite{zhou2024safedrive}, which extends an earlier model of the risk a human driver perceives ahead of the vehicle. The construction has four steps: a kinematic prediction of the path the ego vehicle would follow, a Gaussian field laid along that path, a cost assigned to each agent according to the severity of a potential conflict, and an accumulation of the cost-weighted field over the scenario. The remainder of this appendix presents the four steps in turn; the resulting scalar is the value stored per frame as \texttt{total\_risk} (highD) or \texttt{risk\_value} (the other sources).

\paragraph{Predicted path.}
The ego vehicle is described by a kinematic car model with position $(x,y)$, heading $\phi$ and steering angle $\delta$. Under a constant steering angle, the vehicle follows a circular arc whose radius is
\[
R = \frac{L}{\tan\delta},
\]
where $L$ is the wheelbase. The heading is taken as $\phi = \text{heading}\cdot\pi/180$, and where a source does not log the steering angle directly, $\delta$ is inferred from the change in heading and a steering ratio $S_r$ that maps the steering-wheel angle to the road-wheel angle. The radius $R$ fixes the center of the turning circle $(x_c,y_c)$ from the vehicle position and heading, and the arc length $s$ is measured along the predicted path from the vehicle to a grid point.

\paragraph{Gaussian field along the path.}
The risk contributed by one vehicle at a spatial grid point $(X,Y)$ is modeled as a torus with a Gaussian cross-section centered on the predicted arc:
\[
\mathcal{R}(X,Y) = a(s)\,\exp\!\Bigg(
-\frac{\big((X-x_c)^2 + (Y-y_c)^2 - R^2\big)^2}{2\sigma^2}
\Bigg),
\]
where the height $a(s)$ and the width $\sigma$ both depend on the arc length $s$. The height is largest near the vehicle and decreases along the predicted path; following SafeDrive it is a parabolic function of $s$ whose peak is set by a steepness parameter $p$ and whose reach grows with speed through a look-ahead time $t_{la}$. The width broadens with distance and with steering, and differs between the inner and outer sides of a curve:
\[
\sigma_i = (m + k_i\,|\delta|)\,s + c, \qquad i \in \{1\ (\text{inner}),\, 2\ (\text{outer})\}.
\]
Here $c = w_{\text{car}}/4$ sets the width at the vehicle, so that $\pm 2\sigma$ covers about $95\%$ of the lateral spread, $m$ sets the width during straight driving ($\delta = 0$), and $k_1$ and $k_2$ broaden the inner and outer sides as the steering magnitude $|\delta|$ grows. The asymmetry between $k_1$ and $k_2$ reflects driving styles such as curve-cutting or overshooting. The field is therefore parameterized by $p$, $t_{la}$, $m$, $c$, $k_1$ and $k_2$, and depends only on the ego state $(v_e, \phi, \delta)$, with $v_e = \sqrt{v_x^2 + v_y^2}$.

\paragraph{Event cost and risk accumulation.}
Each agent in the scenario is assigned a cost weight $\alpha$ that reflects the severity of a potential conflict with it, set by the agent class so that large or vulnerable road users such as trucks, buses, pedestrians and cyclists receive higher weights than passenger cars. The cost map is combined with the field by element-wise multiplication and summed over the grid, giving the quantified perceived risk
\[
\mathrm{QPR} = \sum_{\text{grid points}} \big(\alpha \cdot \mathcal{R}\big).
\]

\paragraph{Omnidirectional accumulation.}
A classical DRF covers only the forward half-plane. To assess risk in dense, interaction-rich traffic, the field is extended to a full $360^\circ$ view by adding the contributions of leading and following agents:
\[
\mathrm{QPR}_{\text{total}} = \mathrm{QPR}_{\text{front}} + \mathrm{QPR}_{\text{rear}},
\]
\[
\mathrm{QPR}_{\text{front}} = \sum_{\text{grid points}} \big(\alpha_{\text{front}} \cdot \mathcal{R}_{\text{ego}}\big),
\qquad
\mathrm{QPR}_{\text{rear}} = \sum_{\text{grid points}} \big(\alpha_{\text{ego}} \cdot \mathcal{R}_{\text{rear}}\big),
\]
where $\mathcal{R}_{\text{ego}}$ is the field of the ego vehicle weighted by the cost of the agents ahead, and $\mathcal{R}_{\text{rear}}$ is the field of the following agents weighted by the cost of the ego vehicle. The per-frame scalar $\mathrm{QPR}_{\text{total}}$ is the DRF value to which the annotation protocol applies its top-percentile thresholds.

\section{K-Risk Event Schema}
\label{appendix:schema}

This appendix documents the field-level schema of a K-Risk event. Each event is stored as a synchronized triple, comprising a CSV file of raw trajectory records, a JSON file of per-frame metadata, and a text file of natural language annotation, all sharing the event identifier encoded in the file name. \tabref{tab:schema-event} lists the event-level descriptors carried by the file name and folder, \tabref{tab:schema-traj} the per-frame trajectory fields, and \tabref{tab:schema-text} the components of the natural language annotation. Each event retains the native column names of its source dataset rather than being renamed to a single shared vocabulary, so the per-frame fields are described by their role together with the column names used across sources; fields that are absent in a given source are simply not present in its files.

\begin{table}[!ht]
\centering
\footnotesize
\setlength{\tabcolsep}{5pt}
\renewcommand{\arraystretch}{1.2}
\begin{tabularx}{\linewidth}{@{}p{2.6cm} p{4.6cm} X@{}}
\toprule
\textbf{Field role} & \textbf{Column name(s) across sources} & \textbf{Description (unit)} \\
\midrule
frame index & \col{frame}; \col{frame_id}; \col{Frame_ID} & Frame index within the recording. \\
agent identifier & \col{id}; \col{car_id}; \col{trackId}; \col{Vehicle_ID} & Identifier of the agent in the frame. \\
position & \col{x},\,\col{y}; \col{car_center_x},\,\col{car_center_y}; \col{xCenter},\,\col{yCenter}; \col{Local_X},\,\col{Local_Y} & Planar position of the agent center (m). \\
size & \col{width},\,\col{height}; \col{length}; \col{v_Length},\,\col{v_Width} & Vehicle length and width (m). \\
speed & \col{speed}; \col{v_Vel} & Scalar speed of the agent (m/s). \\
velocity components & \col{xVelocity},\,\col{yVelocity}; \col{vx},\,\col{vy}; \col{lonVelocity},\,\col{latVelocity} & Longitudinal and lateral (or $x$ and $y$) velocity (m/s). \\
acceleration & \col{xAcceleration},\,\col{yAcceleration}; \col{acceleration}; \col{lonAcceleration},\,\col{latAcceleration}; \col{v_Acc} & Longitudinal (and lateral) acceleration (m/s$^2$). \\
heading & \col{heading}; \col{course} & Heading or course angle of the agent (deg). \\
yaw rate & \col{yaw_rate} & Rate of change of heading, CitySim sources (deg/s). \\
lane & \col{laneId}; \col{lane_id}; \col{Lane_ID} & Lane occupied by the agent. \\
agent class & \col{vehicle_class}; \col{class}; \col{v_Class} & Agent class (car, truck/bus, pedestrian, cyclist, \ldots); absent for CitySim. \\
spacing & \col{dhw},\,\col{thw}; \col{Space_Headway},\,\col{Time_Headway} & Distance and time headway to the preceding vehicle (m, s). \\
time-to-collision & \col{ttc}; \col{TTC}; \col{TTC_0_1s},\,\col{TTC_1_2s} & Time-to-collision estimate (s). \\
trajectory conflict & \col{collision_0_1s},\,\col{collision_1_2s},\,\col{collision_ids_0_1s},\,\col{collision_ids_1_2s} & Conflict flags within the next one and two seconds and the conflicting-agent identifiers, CitySim sources. \\
risk value & \col{total_risk}; \col{risk_value} & Driver risk field value for the frame. \\
surrounding agents & \col{precedingId},\,\col{followingId},\,\ldots; \col{preceding_id},\,\col{following_id},\,\ldots & Identifiers of the up to eight surrounding agents by spatial role (preceding, following, left/right-preceding, left/right-following, left/right-alongside). \\
behavioral flags & \col{acc_high},\,\col{brake_high},\,\col{yaw_left},\,\col{yaw_right}; \col{acc_label},\,\col{brake_label},\,\col{left_turn_label},\,\col{right_turn_label} & Boolean flags raised by the annotation protocol; the highD and CitySim sources. \\
\bottomrule
\end{tabularx}
\caption{Per-frame trajectory fields in the CSV file and the JSON per-frame records. Because each event preserves the native column names of its source dataset, the fields are listed by role together with the corresponding names used across highD, CitySim (ExpresswayA, FreewayB), LevelXData (inD, rounD), and NGSIM sources, separated by semicolons. Fields not recorded by a given source are omitted from its event files rather than imputed, thereby preserving the original information without introducing artificial values. The CitySim sources additionally store the four bounding-box vertex coordinates of each vehicle, enabling accurate reconstruction of vehicle orientation, footprint, and spatial occupancy at each frame. This unified field mapping facilitates consistent interpretation of heterogeneous trajectory records while retaining the native data representation of each source dataset.}
\label{tab:schema-traj}
\end{table}

\begin{table}[!ht]
\centering
\footnotesize
\setlength{\tabcolsep}{5pt}
\renewcommand{\arraystretch}{1.2}
\begin{tabularx}{\linewidth}{@{}p{2.6cm} p{2.0cm} X@{}}
\toprule
\textbf{Field} & \textbf{Type} & \textbf{Description} \\
\midrule
\texttt{event\_id} & string & Full event identifier, given by the file name (e.g.\ \texttt{expresswayA\_track\_1\_car\_135\_frame\_640\_to\_735} or \texttt{highd\_01\_113\_precedingId\_yaw\_right\_frame\_2205\_to\_2312}). \\
\texttt{source\_dataset} & string & Upstream source from which the event was extracted, given by the file-name prefix and folder (highd, ind, round, expresswayA, freewayB, ngsim). \\
\texttt{ego\_id} & integer & Identifier of the ego vehicle within the source recording. \\
\texttt{risk\_level} & categorical & Severity grade given by the containing folder: normal (moderate), high or extreme. \\
\texttt{frame\_start}, \texttt{frame\_end} & integer & First and last frame of the extracted event window, spanning the lead-in, peak-risk and resolution phases. \\
\texttt{relation} & categorical & Spatial role of the interacting agent relative to the ego vehicle (preceding, following, left/right-preceding, left/right-following, left/right-alongside); present in the highD file names. \\
\texttt{behavior} & categorical & Triggering behavior of the interaction (acc\_high, brake\_high, yaw\_left, yaw\_right); present in the highD file names. \\
\bottomrule
\end{tabularx}
\caption{Event-level descriptors encoded in the file name and the containing folder of each event. The road structure (highway, urban freeway, intersection or roundabout) follows from \texttt{source\_dataset}, and the peak-risk frame lies inside the event window.}
\label{tab:schema-event}
\end{table}

\begin{table}[!ht]
\centering
\footnotesize
\setlength{\tabcolsep}{5pt}
\renewcommand{\arraystretch}{1.2}
\begin{tabularx}{\linewidth}{@{}p{3.2cm} X@{}}
\toprule
\textbf{Component} & \textbf{Description} \\
\midrule
scenario description & Structured natural language summary of the road layout, lane topology, legal maneuvers, and the dynamic states of the ego vehicle and surrounding agents, generated per frame of the event window. \\
risk reminders & Rule-based notifications attached to the description, keyed by the spatial relation and the triggered behavioral flag, that flag abnormal behavior and the presence of vulnerable road users. \\
risk analysis & LLM-generated explanation of what produced the hazard, why it is safety-critical, how it evolves, and which response is recommended. \\
action\_id & Discrete decision drawn from the five-action schema (1~IDLE, 2~Turn~Left, 3~Turn~Right, 4~Acceleration, 5~Deceleration). \\
\bottomrule
\end{tabularx}
\caption{Components of the natural language annotation stored in the text (JSON Lines) file.}
\label{tab:schema-text}
\end{table}

\newpage

\section{Worked Example of the Closed-Loop LLM Annotation Pipeline}
\label{appendix:closed_loop_walkthrough}

This appendix expands the schematic in \figref{fig:llm_annotation_loop}
into a concrete walk-through of one extreme-risk event from the K-Risk
release. The event is drawn from the ExpresswayA source, in which ego
vehicle~134 travels eastbound on lane~10 of an urban freeway. During the
event window, preceding vehicle~113 in the same lane brakes sharply and
nearly stops, causing the time-to-collision to drop to approximately
$1.6$~s. This rapid closing interaction places the ego vehicle in an
imminent rear-end conflict, and the event is therefore included in the
extreme subset and graded as extreme risk by the annotation protocol.
Every fragment shown below is reproduced from the corresponding event
files in the release, including the trajectory records, textual
description, risk analysis and decision recommendation. Ellipses
($\ldots$) indicate omitted frames or fields that are not essential to
the walk-through.

\begin{figure}[htbp]
\centering
\includegraphics[width=1\linewidth]{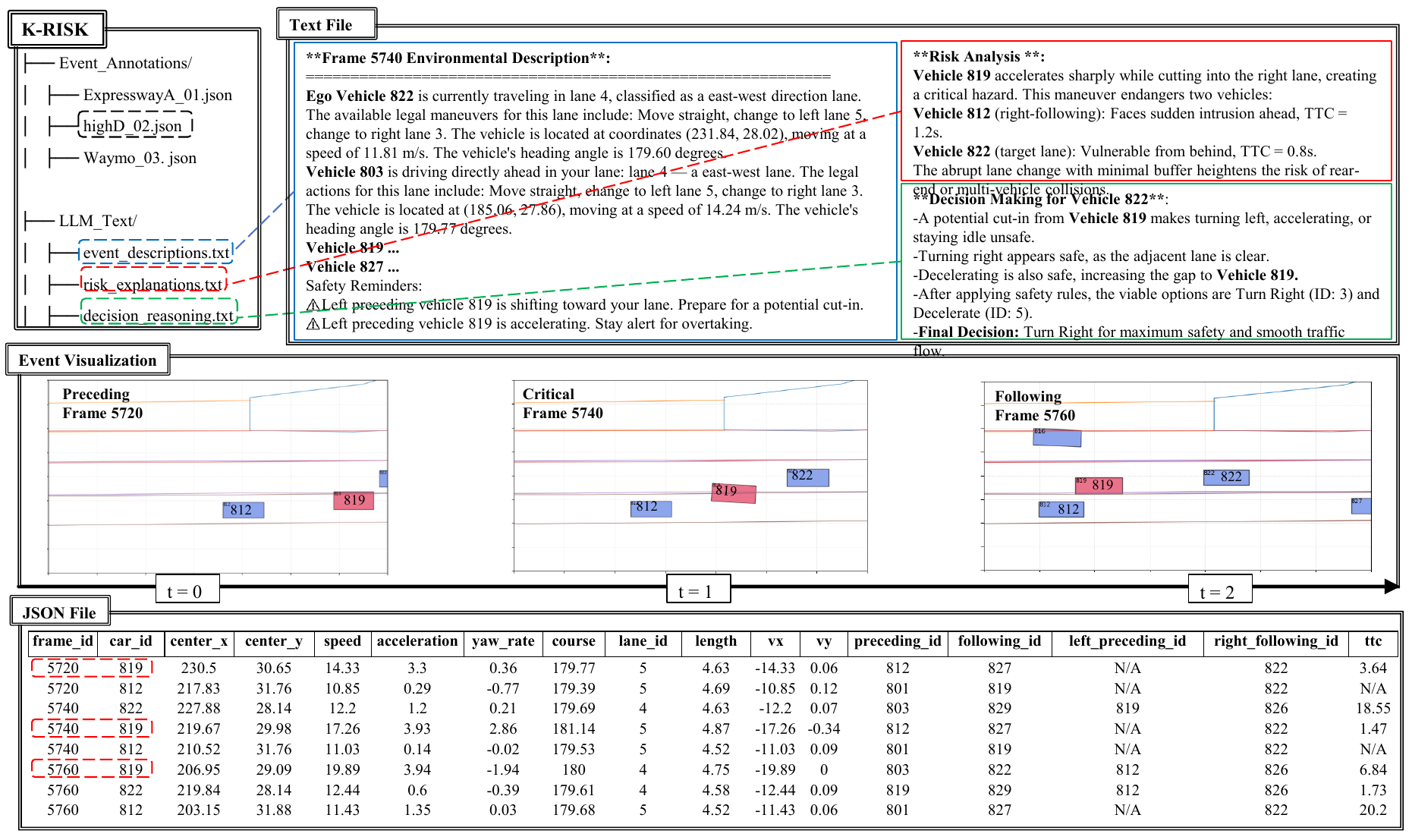}
\caption{Worked example of a single K-Risk event, drawn from the
ExpresswayA source. The text layer holds the environmental description, the
LLM risk analysis and the decision for the ego vehicle (top left).
Time-stamped snapshots show the preceding, critical and resolution
frames of the event window (middle). The JSON layer lists the per-frame
trajectory records, position, speed, acceleration, heading, lane and
the eight surrounding-vehicle role identifiers, for the ego and its
neighbors (bottom).}
\label{fig:walkthrough_records}
\end{figure}

\begin{figure}[p]
\centering
\includegraphics[
    width=\linewidth,
    height=0.80\textheight,
    keepaspectratio
]{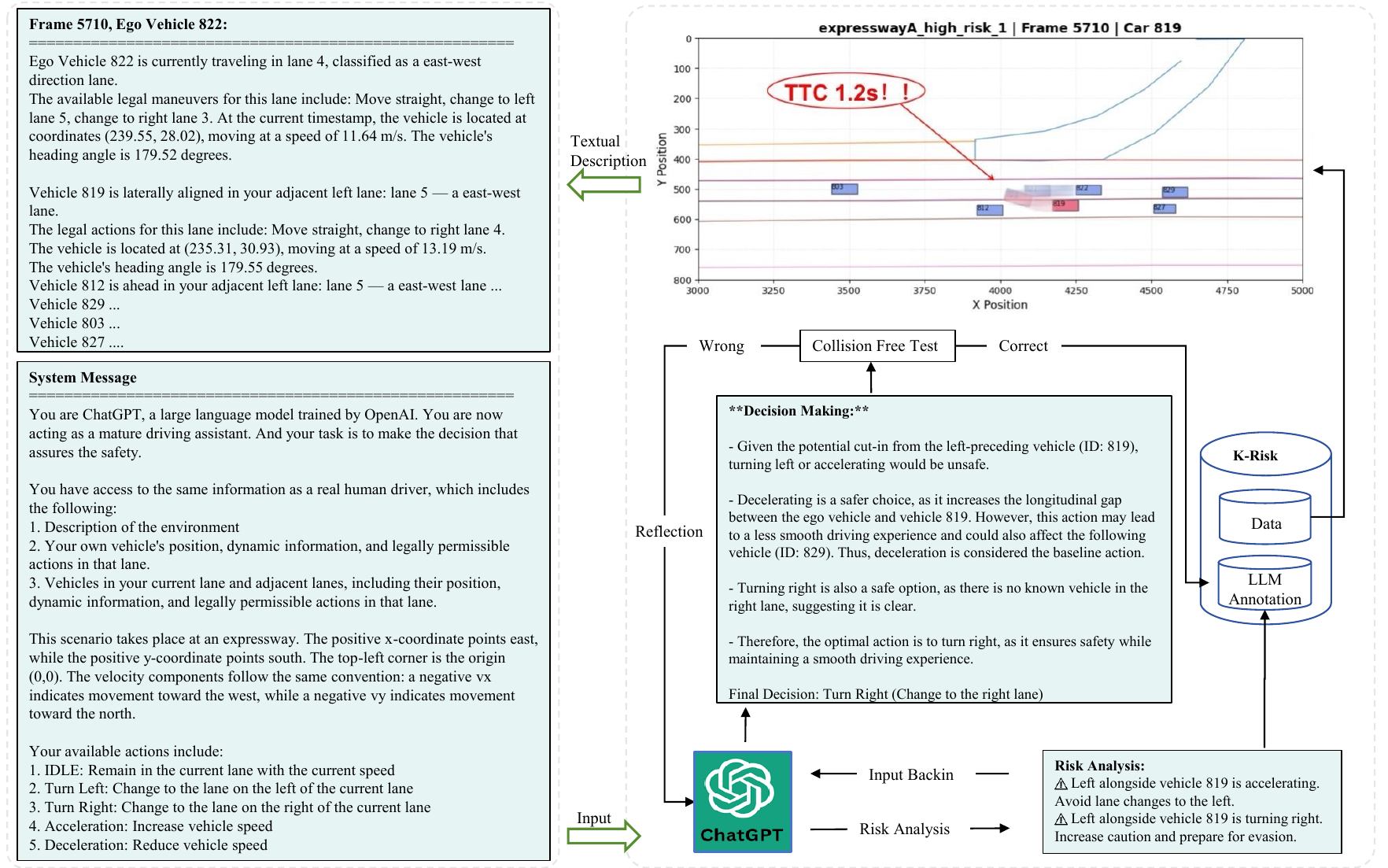}
\caption{Worked example of the closed-loop LLM annotation on an 
ExpresswayA event. The structured scenario description of ego
vehicle~822 and its neighbors, together with the domain-specific system
prompt and its five-action schema, form the LLM input (left). The
trajectory visualization marks the critical frame and the TTC of about
$1.2$~s, with the safe and unsafe action regions indicated (top right).
The LLM produces a risk analysis and a decision rationale and selects a
single action, here a right lane change away from the accelerating
left-adjacent vehicle~819 (right). The chosen action is checked in
simulation, and the reflection step feeds the validated risk analysis
and decision back into the K-Risk release (bottom).}
\label{fig:walkthrough_llm}
\end{figure}

Before the field-by-field walk-through, \figref{fig:walkthrough_records}
and \figref{fig:walkthrough_llm} provide a graphical overview of the same
closed-loop annotation process applied to an ExpresswayA event. In this
example, ego vehicle~822 travels eastbound on lane~4, while the
left-adjacent vehicle~819 accelerates and cuts toward the ego lane,
driving the time-to-collision down to approximately $1.2$~s. This
interaction illustrates how a short temporal window is converted into a
structured high-risk event with aligned textual, visual and trajectory
records.

\figref{fig:walkthrough_records} shows how one such event is organized
into the synchronized triple defined in \figref{fig:event_structure}.
The text layer records the environmental description, the risk analysis
and the recommended decision. The visualization layer presents the
preceding, critical and resolution frames of the event window, making the
temporal evolution of the conflict directly inspectable. The JSON layer
stores the per-frame trajectory records of the ego vehicle and its
surrounding agents, including their positions, speeds, accelerations,
headings and lane information where available.

\figref{fig:walkthrough_llm} traces the corresponding annotation loop
end to end. The loop begins with the structured scenario description and
the system prompt, proceeds through the LLM-based risk analysis and
action selection, and ends with a reflection step that uses validated
outcomes to improve the consistency of the released annotations. The
five stages that follow then examine a distinct ExpresswayA event, in
which ego vehicle~134 encounters the sharply braking preceding
vehicle~113. This case is read field by field to instantiate the same
closed-loop annotation logic on a concrete extreme-risk scenario.

\paragraph{Stage 1: System prompt and action schema.}
The scenario generator prepends a domain-specific system prompt that
fixes the legal-action schema. The five action identifiers in this
prompt are the schema referenced in the Usage Notes and in
\tabref{tab:downstream}.

\begin{lstlisting}[style=walkthrough]
You are ChatGPT, ..., now acting as a mature driving assistant. Your task
is to make the decision that assures the safety. You have access to the
same information as a real human driver, ... This scenario takes place at
an expressway with exits.

Your available actions include:
1. IDLE:         Remain in the current lane with the current speed   (Action ID: 1)
2. Turn Left:    Change to the lane on the left of the current lane  (Action ID: 2)
3. Turn Right:   Change to the lane on the right of the current lane (Action ID: 3)
4. Acceleration: Increase vehicle speed                               (Action ID: 4)
5. Deceleration: Reduce vehicle speed                                 (Action ID: 5)

A clear action ID must be chosen at the end of your reasoning process.
\end{lstlisting}

\paragraph{Stage 2: Trajectory record (CSV $\rightarrow$ JSON event slice).}
The raw labeled trajectory stores per-frame kinematics together with the
eight-surrounding-vehicle role columns (\texttt{preceding\_id},
\texttt{following\_id}, \texttt{left\_preceding\_id}, $\ldots$,
\texttt{right\_following\_id}), the behavioral flags
(\texttt{acc\_label}, \texttt{brake\_label}, \texttt{left\_turn\_label},
\texttt{right\_turn\_label}) and the conflict fields
(\texttt{collision\_in\_2s}, \texttt{Time to Collision}) used by the
annotation protocol. The event-extraction step slices this trajectory
into a list of per-frame, per-agent records and writes the JSON file
shown below; only the peak-risk frame~769 for ego~134 and a subset of
its surrounding agents are reproduced here.

\begin{lstlisting}[style=walkthrough]
[
  { "car_id": 134, "frame_id": 769,
    "car_center_x": 134.71, "car_center_y": 55.94,
    "speed": 7.08, "course": 0.32, "lane_id": 10,
    "vx": 7.08, "vy": 0.04, "acceleration": -0.38, "yaw_rate": -0.21,
    "preceding_id": 113, "following_id": null,
    "left_preceding_id": 131, "left_alongside_id": 133, "left_following_id": 137,
    "right_preceding_id": null,
    "acc_label": false, "brake_label": false,
    "left_turn_label": false, "right_turn_label": false,
    "collision_in_2s": true, "collision_id": 113, "Time to Collision": 1.6 },
  { "car_id": 113, "frame_id": 769,
    "car_center_x": 149.22, ..., "speed": 0.03,
    "acceleration": -5.17, "brake_label": true,
    "collision_in_2s": true, "collision_id": 134, "Time to Collision": 1.45 },
  { "car_id": 131, "frame_id": 769, ... },
  { "car_id": 133, "frame_id": 769, ... },
  ...   /* frames 679, 684, ..., 799 follow */
]
\end{lstlisting}

\paragraph{Stage 3: Structured scenario description and rule-based risk reminders.}
The generator converts each retained frame into the structured English
description that is later fed to the LLM. The \texttt{Safety Reminder}
lines are produced from the rule-based risk filters of the annotation
protocol, keyed by the spatial relation and the triggered condition,
here a short time-to-collision and a sharply decelerating preceding
vehicle. The excerpt below shows the peak-risk frame at
$\mathrm{frame} = 769$, when vehicle~113 has almost stopped directly
ahead of the ego vehicle.

\begin{lstlisting}[style=walkthrough]
Frame 769:
    Ego Vehicle 134 is currently traveling in lane 10, a west-east lane.
    The available legal maneuvers for this lane include:
        Move straight to lane 11, change to left lane 8.
    - Position: (134.71, 55.94) m
    - Speed: 7.08 m/s,  Heading: 0.32 degrees

    Vehicle 113 is driving directly ahead in your lane (lane 10).
        - Position: (149.22, 55.49) m,  Speed: 0.03 m/s
    Vehicle 131 is ahead in your adjacent left lane (lane 8): 4.33 m/s.
    Vehicle 133 is laterally aligned in your adjacent left lane (lane 8): 5.01 m/s.

Safety Reminders:
- Preceding vehicle 113 is decelerating sharply.
  Immediately decelerate to maintain a safe distance.
- [TTC Risk Assessment] Vehicle 113 poses a collision risk within 1-2 seconds.
  Estimated TTC: 1.6.
\end{lstlisting}

\paragraph{Stage 4: LLM risk analysis and chosen action ID.}
The structured description above is sent to the LLM together with the
system prompt of Stage~1. The response below is reproduced from the
generated annotation. It illustrates the three artefacts that K-Risk
extracts per event: a free-form risk explanation, a decision rationale
grounded in the eight-surrounding-vehicle roles, and a single legal
action ID drawn from the schema of Stage~1.

\begin{lstlisting}[style=walkthrough]
Imminent collision warning:
- Preceding vehicle 113 is directly ahead in the same lane, almost stopped
  (0.03 m/s), while the ego vehicle is moving at 7.08 m/s; TTC is about 1.6 s.
- Any action that maintains or increases speed (IDLE, Acceleration) is unsafe.

Surrounding-vehicle analysis:
- Left-preceding vehicle 131 (lane 8): closing TTC below 5 s, so turning left
  is not allowed.
- Left-alongside vehicle 133 (lane 8): a vehicle alongside on the left blocks a
  left lane change.
- No legal right lane exists in this lane.

After pruning unsafe actions, only deceleration remains. Given the imminent
rear-end risk with a sharply decelerating vehicle ahead, and with both
left-lane options blocked, the only safe and legal maneuver is to decelerate
as quickly and smoothly as possible.

Chosen Action ID: 5  (Deceleration: reduce vehicle speed)
\end{lstlisting}

\paragraph{Stage 5: Closed-loop simulation and reflection.}
The chosen action identifier is forwarded to the constant-acceleration,
constant-steering simulator described in the annotation protocol, which
rolls every relevant agent forward for two seconds under the predicted
action. If any forecast pair of trajectories overlaps within this
two-second horizon, the trial is flagged as a near-collision, the
diagnostic message together with the new state are appended to the
prompt, and the LLM is re-invoked. Iterating this loop three times
produced the first, second and third trial collision rates of $4.58\%$,
$2.67\%$ and $1.91\%$ on the 262 extreme-risk evaluation events
(\tabref{tab:closed-loop-validation}). The first-trial and third-trial
outputs of every such event are persisted side by side, yielding the
preference pairs that the Usage Notes use as direct supervision for
preference optimization without further human labeling.

\makeatletter
\setlength{\@fptop}{0pt}
\setlength{\@fpsep}{8pt plus 1fil}
\setlength{\@fpbot}{0pt plus 1fil}
\makeatother

\end{document}